\documentclass[onecolumn]{article} 
\usepackage[top=3.6cm, bottom=3.2cm, left=2.5cm, right=2.5cm]{geometry}
\usepackage[colorlinks=true,linkcolor=red,citecolor=blue]{hyperref}

\usepackage{amsthm,amsmath,amssymb}
\usepackage{mathrsfs}
\usepackage{upgreek}

\usepackage{tabularx}
\usepackage{graphicx}
\usepackage{multirow}
\usepackage{subfigure}

\usepackage{cite}
\usepackage{url}

\begin{document}
\renewcommand{\thefootnote}{\fnsymbol{footnote}}
\title{Global High Categorical Resolution Land Cover Mapping \\via Weak Supervision\footnote{A website is available at \url{https://x-ytong.github.io/project/PRE-land-cover.html}.}}

\author{Xin-Yi Tong$^1$, Runmin Dong$^2$, Xiao Xiang Zhu$^{1, 3, }$\footnote{Corresponding author.}
\vspace{3mm}
\\
$^1${\em \small Chair of Data Science in Earth Observation, Technical University of Munich, Germany}\\
$^2${\em \small Ministry of Education Key Laboratory for Earth System Modeling,}\\ {\em \small Department of Earth System Science, Tsinghua University, China}\\
$^3${\em \small Munich Center for Machine Learning, Germany}\\}
\date{}
\maketitle

\begin{abstract}
Land cover information is indispensable for advancing the United Nations’ sustainable development goals, and land cover mapping under a more detailed category system would significantly contribute to economic livelihood tracking and environmental degradation measurement. However, the substantial difficulty in acquiring fine-grained training data makes the implementation of this task particularly challenging. Here, we propose to combine fully labeled source domain and weakly labeled target domain for weakly supervised domain adaptation (WSDA). This is beneficial as the utilization of sparse and coarse weak labels can considerably alleviate the labor required for precise and detailed land cover annotation. Specifically, we introduce the Prototype-based pseudo-label Rectification and Expansion (\textit{PRE}) approach, which leverages the prototypes (i.e., the class-wise feature centroids) as the bridge to connect sparse labels and global feature distributions. According to the feature distances to the prototypes, the confidence of pseudo-labels predicted in the unlabeled regions of the target domain is assessed. This confidence is then utilized to guide the dynamic expansion and rectification of pseudo-labels. Based on PRE, we carry out high categorical resolution land cover mapping for 10 cities in different regions around the world, severally using PlanetScope, Gaofen-1, and Sentinel-2 satellite images. In the study areas, we achieve cross-sensor, cross-category, and cross-continent WSDA, with the overall accuracy exceeding $80\%$. The promising results indicate that PRE is capable of reducing the dependency of land cover classification on high-quality annotations, thereby improving label efficiency. We expect our work to enable global fine-grained land cover mapping, which in turn promote Earth observation to provide more precise and thorough information for environmental monitoring.
\end{abstract}

\section{Introduction}
Against the backdrop of global economic development, urbanization increasing, and population growth, the ongoing human planning and reshaping of the land continually impact the Earth’s environment \cite{importance1}. Consequently, the measurement of land cover information becomes critical for informed decision-making related to the United Nations’ sustainable development goals and monitoring the progresses \cite{globalmapping1,importance2}. Accurate and timely large-scale land cover monitoring is urgently required to assist in guiding efforts aimed at improving human livelihoods and mitigating adverse environmental changes, e.g. caused by climate change. For instance, a recent study by Grimes et al. \cite{grimes2024} reports that land cover change in Greenland, driven by significant ice loss and increased temperatures, has led to drastic increases in vegetation and wetlands, and changes in sediment and meltwater distribution, profoundly impacting local geomorphology, albedo, greenhouse gas emissions, and biogeochemical processes.

Satellite remote sensing, characterized by its wide coverage, dynamic observation, and all-weather capabilities, has emerged as an important means for obtaining large-scale land information \cite{mapping,globalmapping2,globalmapping3}. Recent advancements in satellite and machine learning technologies are enabling the automated global land cover mapping at a resolution of 10 m \cite{globalmapping1,esaLC,googleLC}. Using Sentinel satellite imagery, the European Space Agency (ESA) and Google have each released global land cover mapping projects: World Cover \cite{esaLC} and Dynamic World \cite{googleLC}, respectively. Despite offering new opportunities and possibilities for practical global land cover mapping, the scope of these projects is limited to fundamental land cover categories (11 and 9 classes, separately). Under these category systems, some essential land features related to food security \cite{fisheries}, sustainable livelihoods \cite{sustainable}, and well-being \cite{greenspaces}, such as fisheries, industrial zones, roads, and urban green spaces, are challenging to observe. Another notable example is paddy fields, employed for cultivating Asia’s primary staple food, rice, may be classified with a degree of uncertainty as wetlands or water bodies under basic land cover category systems \cite{cropland1,cropland2}. Therefore, there is a crucial need to advance research on large-scale land cover mapping with high categorical resolution, or in other words, a fine-grained category system.

However, achieving fine-grained land cover classification requires classifiers capable of distinguishing the spectral, texture, shape, and spatial distribution information of landscapes at a finer level \cite{challenge,scaleadaptive}. In addressing this challenge, deep learning has demonstrated significant superiority. Deep Convolutional Neural Networks (DCNNs) can adaptively represent the complicated contextual information contained in remote sensing images through multi-layer transformations, enabling the identification of heterogeneous land structures and patterns \cite{deep1,deep2,deep3}. Nevertheless, the performance of DCNNs heavily relies on the quality and quantity of training data \cite{earthnets,dong2021high,dong2020improving}. Obtaining large-scale, dense annotations of land cover is an extremely labor-intensive and time-consuming process \cite{GID,FBP,sslscene}, which also demands relevant expertise in the field of remote sensing to ensure the labeling accuracy. This situation makes fully supervised deep learning impractical for real-world fine-grained land cover mapping.

To alleviate the dependence of DCNNs models on precisely annotated data, unsupervised domain adaptation (UDA) and weakly supervised semantic segmentation (WSSS) have been considered as alternatives in recent remote sensing literature.

UDA aims to transfer knowledge from fully labeled data (referred to as the \textit{source domain}) to unlabeled data (referred to as the \textit{target domain}), thereby generalizing models fitted in a small region to a larger region \cite{GID,FBP}. Recent developments indicate that UDA can effectively bridge the domain distribution gap caused by variations in sensor imaging conditions, such as lighting, season, resolution, etc \cite{udaLoveCS,udaconsistencyre,udaHighDAN}. However, due to the absence of supervision information in the target domain, UDA may struggle to achieve generalizability when confronted with substantial shifts in the feature distribution, as seen in scenarios involving differences in geographical locations and inconsistencies in category systems \cite{xu2023}.

WSSS is dedicated to training semantic segmentation networks with sparse and coarse supervision, where only image-, point-, line-, or block-wise labels are annotated for training data \cite{wssspoint, wsssimage,wsssblock}. By avoiding the precise delineation of complex boundaries and ambiguous pixels, the labor required for land cover annotation can be significantly reduced \cite{wsssFESTA}. Recent studies demonstrate that WSSS has the capability to enhance the performance of segmentation models trained with weak labels \cite{wsssCRGNet,wsssDBFNet,wsssKE-WESUP}. However, since the training samples near the boundaries of ground objects are not provided, WSSS always tends to generate fragmented and blended outcomes in the border regions, making it unfeasible to produce results closely resembling those achieved through dense, pixel-wise supervision.

Given the intractable situations faced by UDA and WSSS, it is natural to ask: \textit{Can weak labels serve as a bridge between the source domain and the target domain to mitigate severe shifts in the domain distribution?} From this perspective, dense annotations in the source domain can offer detailed information of object boundaries, while sparse annotations in the target domain can anchor the feature distributions of ground objects \cite{cvwsda2,cvwsda1}. Is the combination of them a feasible solution?

Building on this thinking and with the goal of achieving global fine-grained land cover mapping at a low annotation cost while ensuring high quality in categorical resolution and segmentation accuracy, we propose the integration of fully labeled source domain with weakly labeled target domain, thus entering the scenario of weakly supervised domain adaptation (WSDA). Inspired by the recent success of land cover classification method based on the prototypes (i.e., the class-wise feature centroids) \cite{SSL}, we present a WSDA approach that links labeled and unlabeled regions in the target domain utilizing the prototypes, which is named Prototype-based pseudo-label Rectification and Expansion (\textit{PRE}). Concretely, in addition to using labeled regions from the source and target domains to assess the domain joint segmentation loss, we introduce a dynamic pseudo-label self-training loss and a dynamic pseudo-label self-rectification loss. The model predictions for the unlabeled regions in the target domain are considered as pseudo-labels. And the class likelihoods of these pseudo-labels are estimated according to the feature distances to the prototypes \cite{proda}, which are generated under the guidance of the source domain. A subset of the most reliable pseudo-labels is used to compute the self-training loss. The remaining ones are rectified based on the feature distances, and the degree of modification is used to calculate the self-rectification loss. It is noteworthy that, with each iteration, the pseudo-labels are dynamically expanded, and the prototypes are dynamically updated. Consequently, as the model transfers to the target domain, both pseudo-labels and prototypes are progressively corrected throughout the training.

Based on our PRE method, we carry out high categorical resolution land cover mapping for 10 cities in different regions around the world. Specifically, we use the Five-Billion-Pixels dataset \cite{FBP} as the source domain, which consists of 150 Gaofen-2 (4 m resolution) images distributed within China and annotated with 24 classes. The target domain comprises two parts: (1) One part includes five cites in China: Beijing, Chengdu, Guangzhou, Shanghai, and Wuhan, severally using PlanetScope (3 m), Gaofen-1 (8 m), and Sentinel-2 (10 m) satellite images. The category system aligns with the source domain and is referenced from the Chinese Criteria \textit{GB/T 210102017}; (2) The other part composes five cities located globally: Berlin in Germany, Melbourne in Australia, Nairobi in Kenya, Sao Paulo in Brazil, and Washington DC in the United States, utilizing PlanetScope satellite images. The category system is referenced from the CORINE land cover project \cite{corineLC}, distinct from that of the source domain. Over the study areas, promising results are achieved solely with low-cost weak labels. To the best of our knowledge, we are the first to investigate WSDA for large-scale land cover mapping.

Our contributions are summarized as follows:

\begin{itemize}
\vspace{-2mm}
\item[-]We propose to integrate fully labeled source domain and weakly labeled target domain for land cover mapping. The devised PRE method is able to bridge the gap between the source and target domains using class prototypes and further guide the dynamic expansion and rectification of pseudo-labels in the target domain, thus alleviating severe domain shifts and optimizing predictions at object boundaries.
\vspace{-2mm}
\item[-]We introduce a weakly supervised land cover classification dataset, consisting of two parts: C-megacities and G-cities. C-megacities are located in the eastern, western, northern, southern, and central regions of China, while G-cities are dispersed across Europe, Africa, Oceania, North America, and South America. This dataset is widely distributed, with rich information and fine-grained weak labels, making it highly advantageous for WSSS and WSDA studies.
\vspace{-2mm}
\item[-]We accomplish cross-sensor, cross-category, and cross-continent land cover mapping with high categorical resolution. The results are evaluated with two strategies: sparse label validation and dense label validation (dense labels in the target domain used only for accuracy assessment). Compared to using either the source domain or the target domain alone, our outcomes achieve improvements of over $10\%$ in mIoU for both C-megacities and G-cities, indicating the potential of our work to advance global fine-grained land cover mapping while improving label efficiency.
\end{itemize}

\section{Related work}
\textbf{Unsupervised domain adaptation.} Remote sensing images are susceptible to variations in the imaging conditions and geographic locations, which can result in drastic shifts in the feature distributions \cite{da1,da2}. Consequently, classification models that are optimally trained on well-labeled regions may not be applicable for interpreting images from other regions \cite{GID}. To address this challenge, a commonly employed approach is UDA, involving transferring knowledge from fully labeled source domain to unlabeled target domain. Two major types of UDA have been studied: distribution alignment and self-training.

Distribution alignment methods include two strategies, i.e., discrepancy-based and adversarial-based. Discrepancy-based strategy minimizes the discrepancy criteria between the source and target domains to reduce their distribution distance \cite{discrepancy1,discrepancy2}, achieved through the utilization of manually designed loss functions. In contrast, adversarial-based strategy, such as high-resolution domain adaptation network (HighDAN) \cite{udaHighDAN} and global-local adversarial learning (GOAL) \cite{udaGOAL}, learns criteria by simultaneously training a feature generator and a domain discriminator to alleviate distribution shifts caused by imaging conditions, spatial resolutions, etc. These methods are based on a key assumption that an appropriate match can be found between the two distributions. However, in real-world situations, the feature dispersion and class imbalances within each domain may further lead to the accumulation of intra-domain variance during the alignment process \cite{negativeAdapt}.

Self-training methods treat the model predictions on unannotated regions as pseudo-labels and use them as supervision to fine-tune the current model \cite{GID,FBP}. As models are gradually biased toward the target domain, the reliability of pseudo-labels continues to improve. The key of this type of approaches is to identify confident pseudo-labels and avoid incorporating noisy information into training. Common methods empirically choose a confidence threshold \cite{threshold1,threshold2} or setting a fixed proportion \cite{proportion1,proportion2} for sample selection, which makes it hard to ensure the accuracy of pseudo-labels assigned to the unknown domain. To tackle this issue, dynamic pseudo-label assignment (DPA) \cite{FBP} proposes to dynamically expand pseudo-labels during training iterations, thereby correcting domain shifts in a soft way. Cross-Sensor Land-cOVEr framework (LoveCS) \cite{udaLoveCS} suggests combining adversarial training and self-training to reduce the feature divergence brought by the domain distribution inconsistency, thus further enhancing the performance of UDA.

Despite these commendable efforts, UDA struggles to achieve results comparable to supervised learning, primarily due to the complete absence of supervision information in the target domain, especially when significant distribution differences exist between the two domains.

\textbf{Weakly supervised semantic segmentation.} The largest constraint on the development of large-scale land cover mapping is the scarcity of high-quality pixel-wise ground data. For this issue, a highly promising solution is WSSS, where training data is coarsely annotated with image-, point-, line-, or block-wise labels, significantly reducing the labor and time costs associated with land cover annotating \cite{wssspoint,wsssimage,wsssblock,wsssFESTA}. In this scenario, each image can be divided into labeled and unlabeled regions. The labeled regions can receive direct supervision from the ground truth, while how to learn from the unlabeled regions is the core question in WSSS \cite{tel}. Existing options in current literature fall into two types: consistency regularization and self-training.

Consistency regularization methods, like FEature and Spatial relaTional regulArization (FESTA) \cite{wsssFESTA} and deep bilateral filtering network (DBFNet) \cite{wsssDBFNet}, propagate semantic information from the labeled pixels to the unlabeled pixels by leveraging the internal priors of the image, namely, cross-pixel similarity. Concretely, cross-pixel similarity is established based on spatial and feature relationships among pixels. And a consistency loss is employed to promote the proximity of nearby features, while widening the distance for dissimilar features. Nevertheless, in complex real-world situations, such as with a more detailed category system or larger intra-data variations, these designed consistency criteria may not effectively convey precise semantic information.

Self-training for WSSS aims to gradually expand the sparse annotations with pseudo-labels according to the semantic similarity relationships between pixels. For instance, co-occurrence matrix (CM) \cite{wsssCM} is employed as a label filtering strategy to determine the appropriateness of assigned pseudo-labels. Recent works propose combining consistency regularization and self-training to control the quality of the label expansion mechanism, thus avoiding the introduction of noisy information. Examples include knowledge evolution weakly supervised learning network (KE-WESUP) \cite{wsssKE-WESUP} and consistency-regularized region-growing network (CRGNet) \cite{wsssCRGNet}. However, current self-training strategies focus on expanding pseudo-labels in neighboring pixels, making it hard to comprehensively observe the entire feature distribution. This limitation is likely to result in inaccurate label growing, even under the constraints of consistency criteria.

Although the above efforts have remarkably contributed to reducing the reliance of semantic segmentation models on high-quality annotations, WSSS tends to predict maps with inferior geometric boundaries due to the lack of supervision signals at the object borders. This considerably restricts the practical application of weak labels in land cover mapping.

\begin{figure*}[htb!]
\centering
\includegraphics[width=0.9\textwidth]
{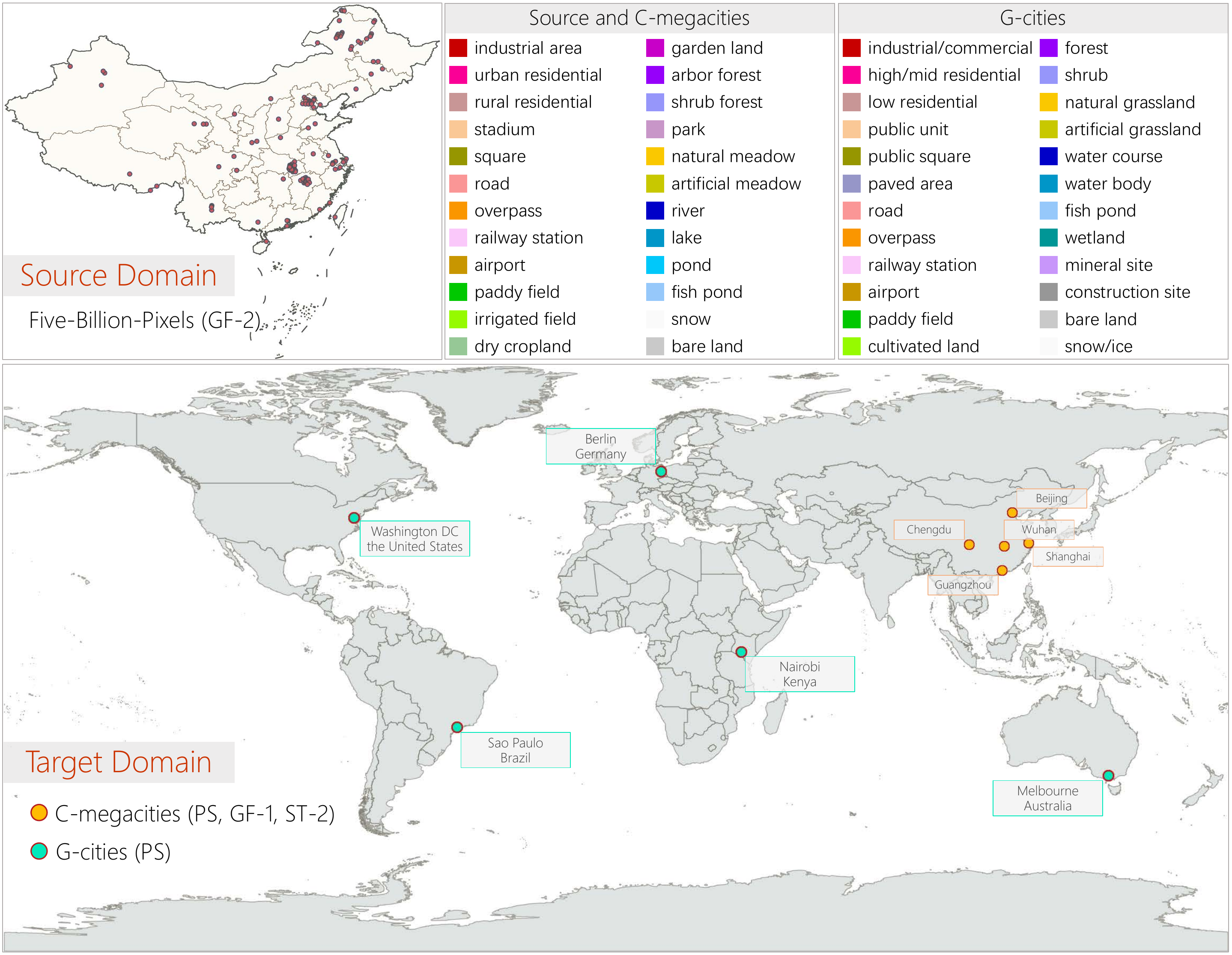}
\caption{The geographical distribution and category systems of C-megacities, G-cities, and Five-Billion-Pixels. The classes of C-megacities are identical to those of the source domain, while the classes of G-cities has been slightly adjusted. Therefore, between G-cities and the source domain, there exists cross-sensor, cross-category, and cross-continent challenges.}
\label{figure:dataarea}
\end{figure*}

\section{Study area and data}
\subsection{Overview of data}
To reduce the dependence of fine-grained land cover mapping on high-quality dense annotations and to advance the development of low-cost, high-precision semantic segmentation for remote sensing images, we construct a weakly supervised land cover classification dataset, consisting of two parts: C-megacities and G-cities, of which the geographical distribution and category systems are illustrated in Fig. \ref{figure:dataarea}. Using them as the target domain and combining with the source domain, i.e., the Five-Billion-Pixels dataset \cite{FBP}, we implement land cover mapping for 10 cities located globally via WSDA. The introduction of the source and target domains is provided below:

\textbf{Five-Billion-Pixels.} It comprises 150 Gaofen-2 satellite images with a spatial resolution of 4 m, covering 24 classes. The category system is determined with reference to the Chinese Criteria \textit{GB/T 210102017} and is adjusted based on the recognition capabilities of 4 m-resolution optical satellite images. The data source is captured over more than 60 administrative regions in China, encompassing diverse climate, altitude, and geology. The proportion of samples in different categories closely matches the actual distribution of different land covers. Five-Billion-Pixels possesses advantages such as large coverage, wide distribution, and high heterogeneity, opening the door for large-scale land cover mapping under fine-grained category system.

\textbf{C-megacities.} It includes five megacities: Beijing, Chengdu, Guangzhou, Shanghai, and Wuhan, located in the northern, western, southern, eastern, and central regions of China, respectively. The category system is the same as that of Five-Billion-Pixels, referencing from the Chinese Criteria \textit{GB/T 210102017}. The data sources for different megacities are as follows: PlanetScope (PS) for Chengdu and Shanghai, Gaofen-1 (GF-1) for Wuhan, and Sentinel-2 (ST-2) for Beijing and Guangzhou, with the number of images being 8, 8, 4, 1, and 1, respectively. The acquisition times for the images are as follows: for Beijing and Guangzhou in 2021, for Chengdu and Shanghai in 2019, and for Wuhan in 2016. These Chinese megacities are with different geographical environments, development levels, and urban structures. The data sources are captured at diverse spatial resolutions and seasons. Therefore, there are significant variations not only between data domains but also within each domain, which is important for validating the generalizability of the classifiers.

\textbf{G-cities.} It includes five cities: Berlin in Germany, Melbourne in Australia, Nairobi in Kenya, Sao Paulo in Brazil, and Washington DC in the United States, located in Europe, Oceania, Africa, South America, and North America, respectively. The category system covers 24 classes and is referenced from the CORINE land cover project \cite{corineLC}, with some adjustments compared to that of Five-Billion-Pixels. The data source utilized for all cities is PS imagery, with 8 images for each city, making a total of 40 images. The acquisition times range from February 2019 to December 2019. This set of data is situated on different continents, has different categories, and uses different data sources compared to the source domain. It encompasses distinct cultural backgrounds, development levels, urban planning styles, and landscape distributions, posing significant challenges for domain adaptation. It is noteworthy that in this context, ``city'' specifically denotes the location; the study areas actually cover the surrounding rural areas, suburbs, wooded areas, and mountainous regions outside the administrative boundary. Therefore, these study areas can test the performance of algorithms for almost all types of landscapes.

\subsection{Category system}
The category system for C-megacities comprises: \textit{industrial area}, \textit{urban residential}, \textit{rural residential}, \textit{stadium}, \textit{square}, \textit{road}, \textit{overpass}, \textit{railway station}, \textit{airport}, \textit{paddy field}, \textit{irrigated field}, \textit{dry cropland}, \textit{garden land}, \textit{arbor forest}, \textit{shrub forest}, \textit{park}, \textit{natural meadow}, \textit{artificial meadow}, \textit{river}, \textit{lake}, \textit{pond}, \textit{fish pond}, \textit{snow}, \textit{bare land}. As the images are collected in urban and surrounding areas, the category system covers all land types except \textit{tundra} and \textit{permanent ice}. And the ``snow'' category refers to temporary snow and ice.

In the subsequent tables, the abbreviations for classes in C-megacities are defined for convenience as: Indu - industrial area, Ures - urban residential, Rres - rural residential, Stad - stadium, Squa - square, Road - road, Over - overpass, Rail - railway station, Airp - airport, Padd - paddy field, Irri - irrigated field, Dryc - dry cropland, Gard - garden land, Arbo - arbor forest, Shru - shrub forest, Park - park, Nmea - natural meadow, Amea - artificial meadow, Rive - river, Lake - lake, Pond - pond, Fish - fish pond, Snow - snow, Bare - bare land.

Different from C-megacities, the category system for G-cities is referenced from the CORINE land cover project. The CORINE (Coordination of Information on the Environment) project \cite{corineLC} is a flagship component of the European Environment Agency’s (EEA) Copernicus Land Monitoring Service, where it has provided essential information on pan-European land cover inventory for over three decades. Compared to the Chinese Criteria \textit{GB/T 210102017}, CORINE’s category system is more globally applicable. To illustrate, due to China's high population density, the economic development of a region and the distinction between urban and rural residential areas can be inferred through the height and crowdedness of buildings. However, in other parts of the world, low-rise residential areas may also be located in affluent urban regions. For another instance, in China's classification standard, mangroves and forested swamps are categorized as tree forest, marsh grassland is categorized as meadow, and other types of marshes are categorized as water bodies. However, in a more universally applicable classification, all of these should be considered wetlands.

The category system for G-cities comprises: \textit{industrial/commercial}, \textit{high/mid residential}, \textit{low residential}, \textit{public unit}, \textit{public square}, \textit{paved area}, \textit{road}, \textit{overpass}, \textit{railway station}, \textit{airport}, \textit{paddy field}, \textit{cultivated land}, \textit{forest}, \textit{shrub}, \textit{natural grassland}, \textit{artificial grassland}, \textit{water course}, \textit{water body}, \textit{fish pond}, \textit{wetland}, \textit{mineral site}, \textit{construction site}, \textit{bare land}, \textit{snow/ice}. Similarly, this category system encompasses all land types except \textit{tundra} and \textit{permanent ice}. To transfer knowledge from the source domain to G-cities through semantic segmentation, we merge some classes in the source domain and roughly correspond the adjusted categories to those of G-cities, as detailed in Section \ref{sec:setup}.

In the subsequent tables, the abbreviations for classes in G-cities are defined for convenience as: Inco - industrial/commercial, Hres - high/mid residential, Lres - low residential, Puni - public unit, Psqu - public square, Pave - paved area, Road - road, Over - overpass, Rail - railway station, Airp - airport, Padd - paddy field, Cult - cultivated land, Fore - forest, Shru - shrub, Ngra - natural grassland, Agra - artificial grassland, Wcou - water course, Wbod - water body, Fish - fish pond, Wetl - wetland, Mine - mineral site, Cons - construction site, Bare - bare land, Snic - snow/ice.

\subsection{Data source}
\textbf{PlanetScope.} PS is a satellite constellation operated by American Planet Lab, consisting of about 130 individual CubeSats. The sensors capture multispectral (MS) images in the blue, green, red, and near-infrared bands, with a spatial resolution of 3.7-4.1 m, which is resampled to approximately 3 m during data release. The image resolutions of the study data range from $8135\times 3922$ pixels to $10104\times 7875$ pixels.

\textbf{Gaofen-1.} GF-1 is the first satellite of China's High-Definition Earth Observation System (HDEOS). It is equipped with two panchromatic (pan) and MS sensors, providing spatial resolutions of 2 m pan/8 m MS, with a combined swath of over 60 km. The MS images used in this study cover the spectral range of blue, green, red, and near-infrared, with the image resolutions ranging from $4548\times 4503$ pixels to $4548\times 4544$ pixels.

\textbf{Sentinel-2.} ST-2 is a satellite mission within the framework of the European Union’s Copernicus Programme. It currently consists of a constellation of two satellites, Sentinel-2A and Sentinel-2B, providing 13 spectral bands and a field of view of 290 km. In this study, the blue, green, red, and near-infrared bands with a resolution of 10 m are utilized, and the image resolution is $10980\times 10980$ pixels. All the collected images are with very low level of cloud contamination.

\begin{figure*}[htb!]
\centering
\includegraphics[width=0.8\textwidth]
{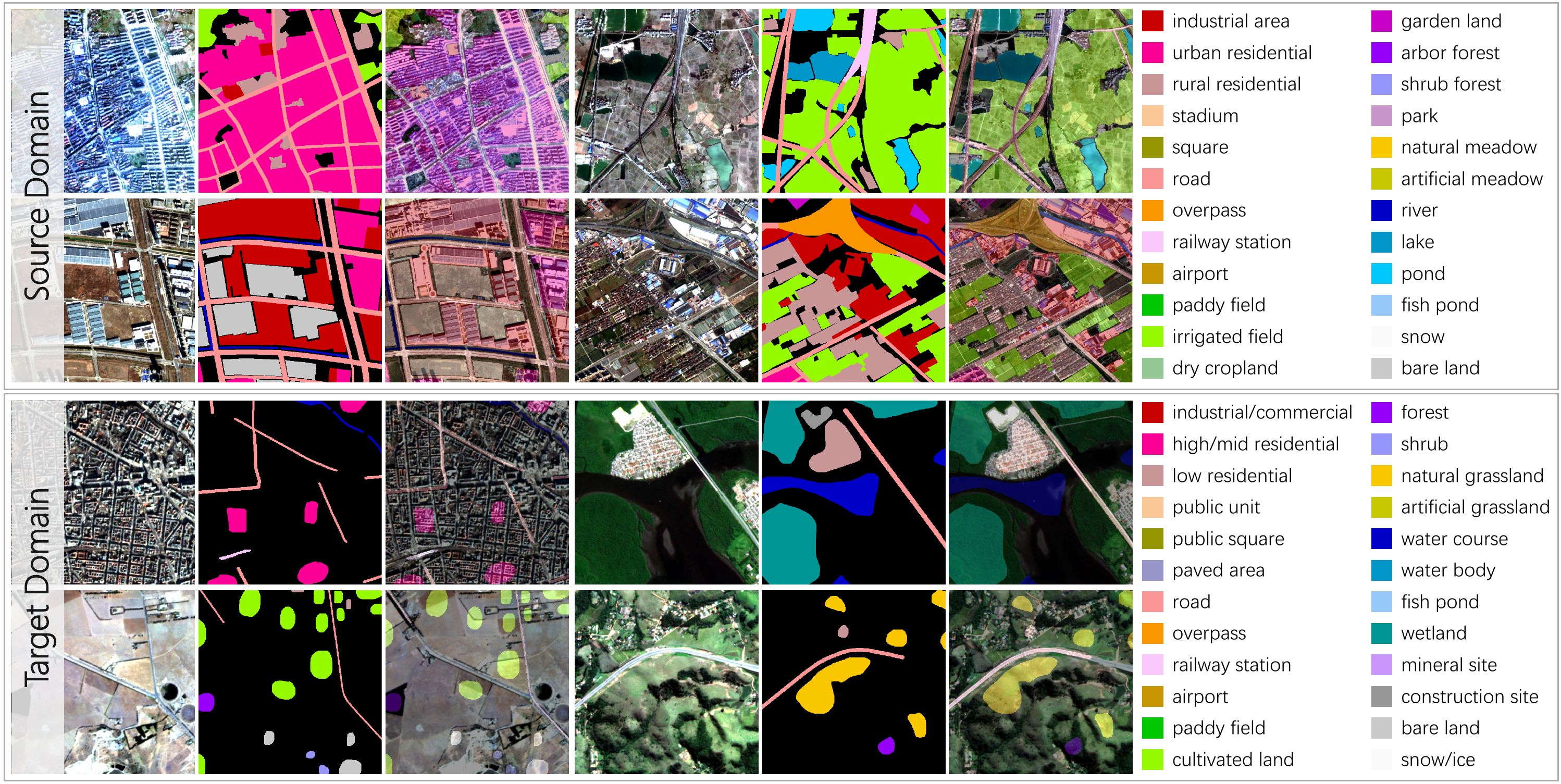}
\caption{Examples of densely annotated source domain and sparsely annotated target domain. Fine delineation for a single $1000\times 1000$-pixel image with a resolution of 3 m takes approximately 1 hour. In contrast, scribbling an image with the same size and resolution takes only about 1 minute due to the avoidance of outlining boundaries.}
\label{figure:datalabel}
\end{figure*}

\subsection{Dataset creation}
We draw weak labels according to the following rules: (1) Professional annotators use the Pencil tool in Adobe Photoshop to rapidly scribble closed areas on the image, and then fill the scribbles with the Paint Bucket tool to create block-wise labels. Due to the need for quick annotation, the borders of the blocks do not align with those of the ground objects (besides roads, we use discontinuous line segments to annotate this category). And these sparse blocks are expected to be distributed evenly across the entire image; (2) Different categories are annotated with different colors, and each block must cover only a single category, excluding pixels from other classes. Although weak labels are incomplete, they need to be accurate. Therefore, after annotation, there is a checking phase where annotators inspect blocks for errors or overflow and make modifications to ensure the reliability of sparse blocks.

Based on our experience, fully and finely annotating a single $1000\times 1000$-pixel image with a resolution of 3 m takes approximately 1 hour. If the spatial resolution is lower, it may take longer as the ground objects become more intricate. In contrast, the sparse and rough annotation of an image with the same size and resolution takes only about 1 minute, demonstrating substantial savings in labor and time costs associated with the creation of weak labels. Examples of densely annotated source domain and sparsely annotated target domain are depicted in Fig. \ref{figure:datalabel}, showing the disparity in their level of delineation.

In total, the sparse labels of C-megacities cover $6.48\times 10^{7}$ pixels, while G-cities is sparsely annotated with a total of $4.11\times 10^{8}$ pixels. The proportions of pixels belonging to each category are listed in Table \ref{table:percentage}.

\begin{table*}[htb!]
\centering
\caption{The percentage of the number of pixels belonging to each category. As can be seen, the category distributions of both C-megacities and G-cities are quite imbalanced.}
\vspace{3mm}
\resizebox{0.8\textwidth}{!}{
\begin{tabular}{lllllllllllll} 
\hline
\textbf{C-megacities}        &       &      &      &      &      &       &       &      &      &      &       &       \\
\hline
Category                     & Indu  & Ures  & Rres & Stad & Squa & Road  & Over  & Rail & Airp & Padd & Irri  & Dryc  \\
Percent (\%)                 & 13.67 & 14.21 & 9.19 & 0.25 & 0.07 & 15.26 & 1.55  & 0.53 & 1.74 & 2.83 & 16.48 & 0.05  \\
\cline{1-1}
Category                     & Gard  & Arbo  & Shru & Park & Nmea & Amea  & Rive  & Lake & Pond & Fish & Snow  & Bare  \\
Percent (\%)                 & 2.28  & 4.43  & 0.24 & 0.40 & 0.88 & 1.45  & 6.39  & 4.81 & 0.32 & 0.44 & 0 	   & 2.54  \\
\hline
\textbf{G-cities}            &       &      &      &      &      &       &       &      &      &      &       &       \\
\hline
Category                     & Inco  & Hres & Lres  & Puni & Psqu & Pave  & Road  & Over & Rail & Airp & Padd  & Cult  \\
Percent (\%)                 & 3.72  & 1.52 & 16.90 & 0.03 & 0.04 & 0.14  & 4.01  & 0.89 & 0.17 & 1.23 & 0.02  & 14.84 \\
\cline{1-1}
Category                     & Fore  & Shru & Ngra  & Agra & Wcou & Wbod  & Fish  & Wetl & Mine & Cons & Bare  & Snic  \\
Percent (\%)                 & 25.49 & 2.32 & 4.04  & 1.51 & 1.57 & 5.85  & 0.05  & 2.28 & 1.04 & 0.34 & 12.00 & 0     \\
\hline
\end{tabular}}
\label{table:percentage}
\end{table*}

\subsection{Data for test}
\begin{figure*}[htb!]
\centering
\includegraphics[width=0.8\textwidth]
{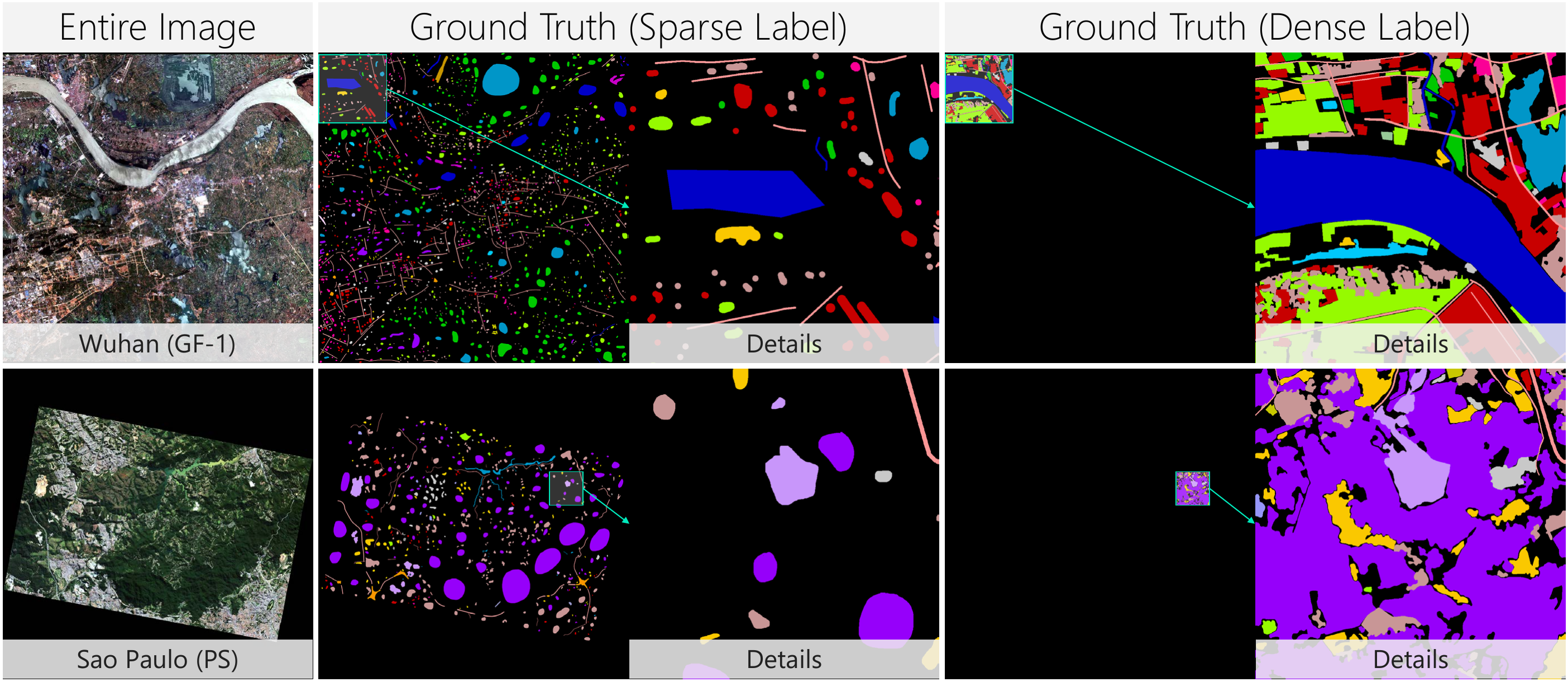}
\caption{Two strategies for quantitative evaluation. Sparse label: coarse blocks are evenly labeled throughout the entire test image. Dense label: sub-regions with sizes of $1000\times 1000$ pixels are labeled on each test image for every city.}
\label{figure:datatest}
\end{figure*}

In C-megacities and G-cities, 3/4 of the images are used for training, and the remaining 1/4 are used for test. Training and test areas have no geographically overlap. To assess accuracy on test images from multiple dimensions, we employ two verification strategies: sparse label-based and dense label-based. Specifically, sparse labeling follows the description mentioned earlier, while dense labeling requires fully and finely annotating $1000\times 1000$-pixel sub-regions on each test image for every city. Sparse labels are used to evaluate the overall classification performance on the entire image, while dense labels can verify the boundary fidelity of the segmentation results in localized regions, as demonstrated in Fig. \ref{figure:datatest}. For C-megacities, the sparse annotations used for testing consist of a total of $1.98\times 10^{7}$ pixels, while the dense annotations comprise $5.34\times 10^{6}$ pixels. 
For G-cities, the sparsely annotated pixels used for testing amount to $1.03\times 10^{8}$, and the densely annotated pixels amount to $7.71\times 10^{6}$. It is important to note that dense annotations in the target domain are only used for accuracy assessment and do not involve model training.

\section{Methodology}
\subsection{Overview of PRE}
\label{sec:overview}
The objective of the WSDA task is to train a semantic segmentation model using fully annotated source data and weakly annotated target data, ensuring effective adaptation of the model to the target domain. We refer to two domains, the source domain $\mathcal{D}_{s}=(\mathcal{X}_{s},\mathcal{Y}_{s})$ and the target domain $\mathcal{D}_{t}=(\mathcal{X}_{t},\mathcal{Y}_{t})$. ${x}_{s}\in\mathcal{X}_{s}$ denotes a source image with the dense label mask ${y}_{s}\in\mathcal{Y}_{s}$, and ${x}_{t}\in\mathcal{X}_{t}$ denotes a target image with the sparse label mask ${y}_{t}\in\mathcal{Y}_{t}$, which is in the form of sparse blocks in this study. ${x}_{t}$ can be divided into two parts, i.e., the labeled area $\Omega_{l}$ and the unlabeled area $\Omega_{u}$, and how to leverage the information from $\Omega_{u}$ is the central challenge that WSDA needs to address.

In UDA and WSSS tasks, an intuitive scheme is self-training \cite{proda,tel}, where the model's predictions on pixels in $\Omega_{u}$ are used as pseudo-labels and incorporated into subsequent model training. However, the model's predictions for the unseen domain are likely to be inaccurate \cite{GID}. To alleviate this issue, in the UDA scenario, of which $\mathcal{D}_{t}$ is unlabeled, reliable pseudo-labels are typically chosen by setting a threshold \cite{threshold1,threshold2} or a ranking ratio \cite{proportion1,proportion2} on prediction confidences. While in the WSSS scenario, the selection of high-quality pseudo-labels often involves examining and expanding the neighboring pixels of $\Omega_{l}$ \cite{wsssFESTA,wsssCRGNet,wsssDBFNet}. Applying these methods to WSDA may impose two limitations: first, due to domain gaps, the model is likely to focus on target samples lying closer to the source distribution; second, the global view in both image space and feature space of the target image is not adequately considered. Consequently, pixels assigned pseudo-labels may lack sufficient information to guide model transfer.

\begin{figure*}[htb!]
\centering
\includegraphics[width=0.8\textwidth]
{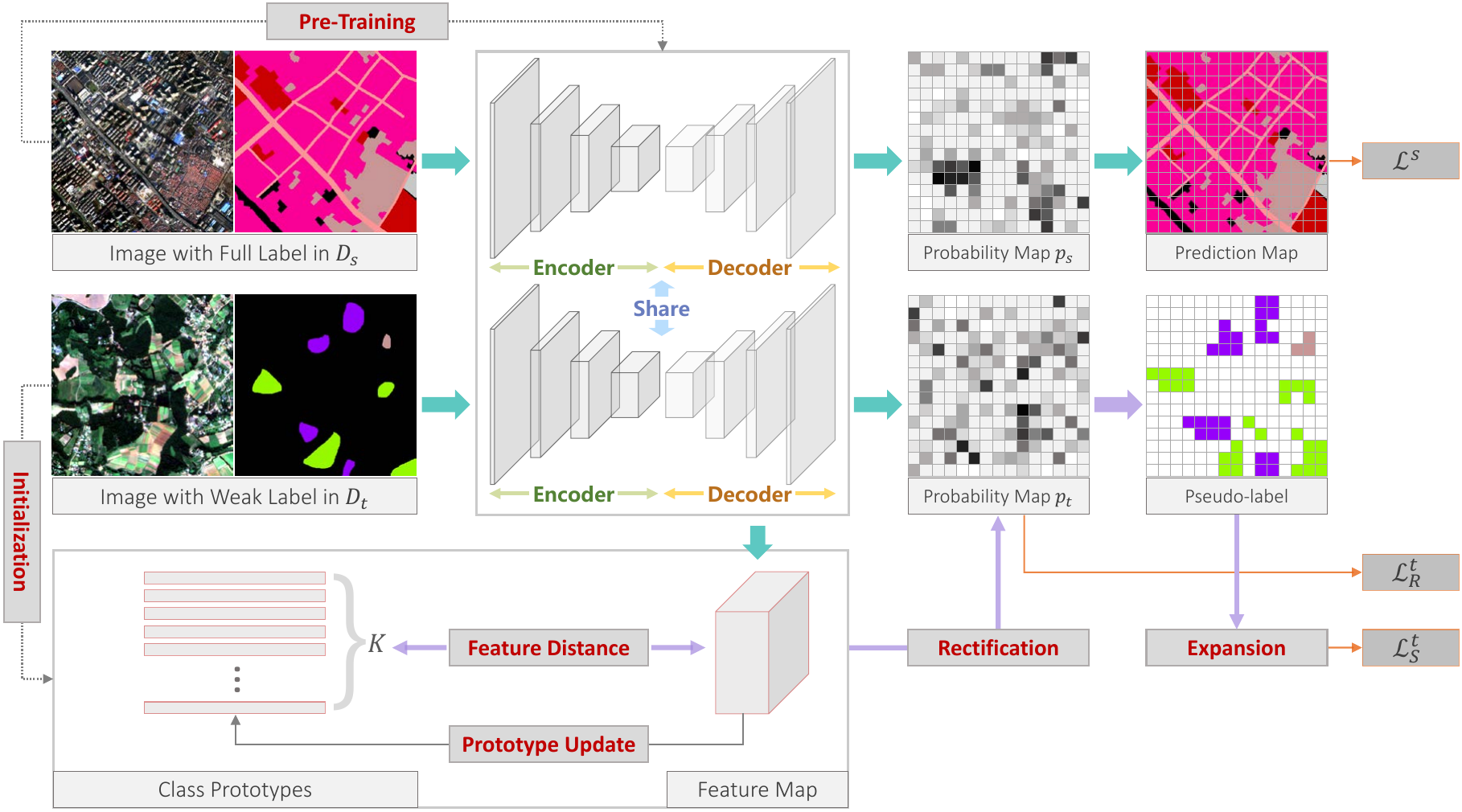}
\caption{Approach overview. We construct a dual-branch semantic segmentation model separately process $\mathcal{D}_{s}$ and $\mathcal{D}_{t}$, with both branches sharing the exact same architecture and parameters. The backbone of each branch is pre-trained on $\mathcal{D}_{s}$. The feature maps of $\mathcal{D}_{t}$ is used to initialize and update the prototypes, while the distances between prototypes and feature maps are used to refine and filter pseudo-labels. Throughout the iterations, pseudo-labels are continuously expanded, and prototypes are dynamically computed. The overall loss function consists of the domain joint segmentation loss, and the self-training loss coupled with the self-rectification loss in the target branch.}
\label{figure:preoverview}
\end{figure*}

To simultaneously leverage information from $\mathcal{D}_{s}$ coupled with $\mathcal{D}_{t}$ and find beneficial pseudo-labels across the entire data distribution of $\mathcal{D}_{t}$, we propose the PRE approach, which consists of four stages: (1) Initialization of the prototypes for $\mathcal{D}_{t}$ (Section \ref{sec:prototype}); (2) Dynamic rectification and expansion of pseudo-labels (Section \ref{sec:pseudolabel}); (3) Model training with class-balanced cross-domain overall loss (Section \ref{sec:lossfunction}); (4) Prototype updating with mini-batch labels in $\mathcal{D}_{t}$ (Section \ref{sec:update}). Stages (2)-(4) are iteratively repeated during the training process. The overall flowchart of PRE is illustrated in Fig. \ref{figure:preoverview}.

\subsection{Class prototype learning}
\label{sec:prototype}
Inspired by the recent success of prototype-based semi-supervised learning for land cover classification \cite{SSL}, we propose adopting prototypes as anchor points to select and rectify pseudo-labels. The prototypes are the averages of class-wise features, calculated using features extracted by a semantic segmentation model pre-trained on $\mathcal{D}_{s}$. Before starting the training of PRE, the model has only learned knowledge from $\mathcal{D}_{s}$, and the goal is to adapt it to the distribution of $\mathcal{D}_{t}$. Therefore, in the early stages of training, the learning of prototypes is mainly guided by $\mathcal{D}_{s}$. As training progresses and the model gradually bias towards $\mathcal{D}_{t}$, both $\mathcal{D}_{s}$ and $\mathcal{D}_{t}$ collectively guide the updating of prototypes.

Specifically, we treat the layer immediately before the classification layer of the segmentation model as the feature extractor, obtaining the feature map ${f}_{t}$ of ${x}_{t}$. The prototype of class ${k}\in\{1,\dots,K\}$ is initialized as

\begin{equation}
\eta^{(k)}=\frac{\sum_{{x}_{t}\in\mathcal{X}_{t}}\sum_{i}{f}_{t}^{(i)}\ast{\mathbb{I}}({y}_{t}^{(i,k)}==1)}{\sum_{{x}_{t}\in\mathcal{X}_{t}}\sum_{i}{\mathbb{I}}({y}_{t}^{(i,k)}==1)},
\end{equation}

where $\mathbb{I}$ is the indicator function, ${f}_{t}^{(i)}$ is the feature vector at the pixel location $i$ on ${x}_{t}$, and ${y}_{t}^{(i,k)}\in\{0,1\}$ is the value of the one-hot label at location $i$ and class $k$.

\subsection{Pseudo-label rectification and expansion}
\label{sec:pseudolabel}
In PRE, the pre-trained semantic segmentation model serves as the backbone, forming a dual-branch model, as presented in Fig. \ref{figure:preoverview}. These two branches share the exact same architecture and parameters, each dedicated to processing $\mathcal{D}_{s}$ and $\mathcal{D}_{t}$. ${p}_{s}$ and ${p}_{t}$ are the prediction maps from the model's classification layer for ${x}_{s}$ and ${x}_{t}$, respectively.

Since samples locating closer to a certain class centroid are more indicative of belonging to that class, the feature distances to the prototypes can be used as the basis for denoising pseudo-labels \cite{proda}. In the target branch, the rectified pseudo-label is calculated as

\begin{equation}
\tilde{y}_{t}^{(i,k)}=\mathcal{P}({w}_{t}^{(i,k)}{p}_{t}^{(i,k)}),
\end{equation}

where $\tilde{y}_{t}^{(i,k)}\in\{0,1\}$ is the value of the one-hot label at index $i$ and class $k$. $\tilde{y}_{t}=\mathcal{P}({p}_{t})$ is the prediction function, defined as

\begin{equation}
\tilde{y}_{t}^{(i,k)}=\begin{cases}1,\ \mathrm{if}\ k=\arg\max_{q}{p}_{t}^{(i,q)}\\0,\ \mathrm{otherwise}\end{cases}
\end{equation}

${w}_{t}^{(i,k)}$ is the weight exploited to adjust the prediction probability, computed as

\begin{equation}
{w}_{t}^{(i,k)}=\frac{\exp(-\left\|{f}_{t}^{(i)}-\eta^{(k)}\right\|)}{\sum_{q}\exp(-\left\|{f}_{t}^{(i)}-\eta^{(q)}\right\|)}.
\end{equation}

\begin{figure*}[htb!]
\centering
\includegraphics[width=0.5\textwidth]
{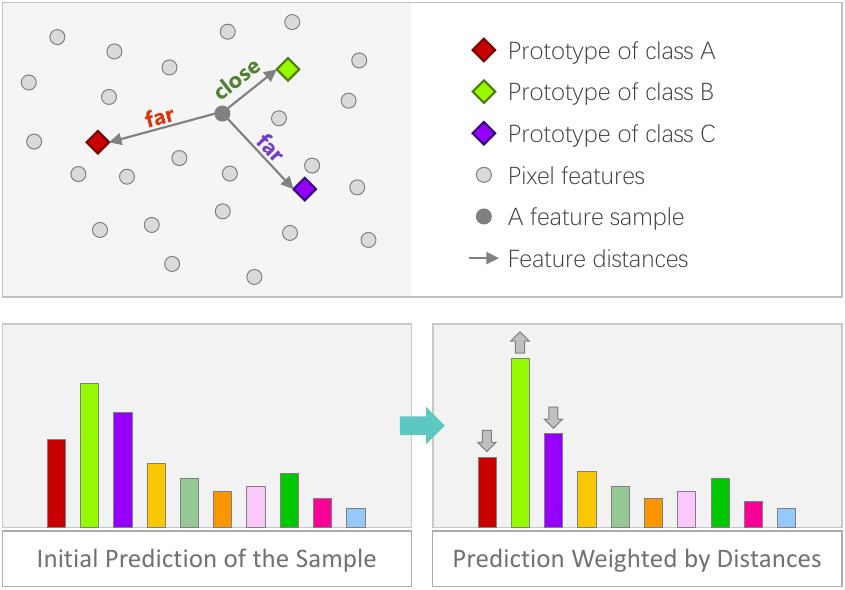}
\caption{The predicted probabilities are rectified based on the feature distances from the prototypes. Categories belonging to prototypes closer to the sample are assigned greater weights, while categories farther from the samples are assigned smaller weights. The weighted probabilities align better with the global feature distribution.}
\label{figure:preprototype}
\end{figure*}

This operation upweights the probability of classifying samples near $\eta^{(k)}$ into category $k$, and conversely downweighs the probability for samples far away from $\eta^{(k)}$ to belong to category $k$, as indicated in Fig. \ref{figure:preprototype}.

Introducing pseudo-labels into self-training with a gradually increasing proportion has proven effective in UDA \cite{FBP}, which is able to reduce the domain shifts and prevent the accumulation of domain variances. In PRE, we adopt this strategy and use the consistency of predicted probabilities before and after rectification to assist in filtering pseudo-labels.

Concretely, for ${x}_{t}$, we arrange the values of ${w}_{t}{p}_{t}$ in descending order and pick out the pixels with the top ${N}$ probability values for self-training. ${N}$ is specified as

\begin{equation}
{N}=\log{(\frac{m}{M}+1)}\sum_{{i}\in\Omega_{u}}{\mathbb{I}}(\mathcal{P}({p}_{t}^{(i)})==\mathcal{P}({w}_{t}^{(i)}{p}_{t}^{(i)})),
\end{equation}

where $m$ is the current epoch count and $M$ is the total number of epochs. The overall set of pixels selected is denoted as $\Omega_{e}$. The expanded sparse label map of ${x}_{t}$ can then be obtained, which is

\begin{equation}
\bar{y}_{t}={y}_{t}\cup\{\tilde{y}_{t}^{(i)}|{i}\in\Omega_{e}\}.
\end{equation}

\begin{figure*}[htb!]
\centering
\includegraphics[width=0.9\textwidth]
{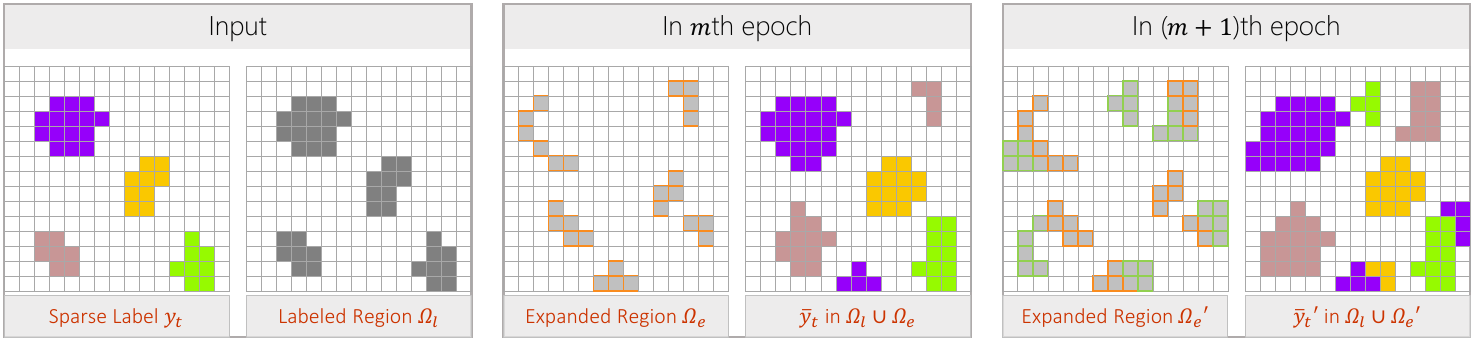}
\caption{The mechanism of dynamic pseudo-label expansion, which enables the model to adapt flexibly to $\mathcal{D}_{t}$ without the need for empirically setting thresholds or weights.}
\label{figure:prepseudolabel}
\end{figure*}

The reason for this design is that when the model is not adapted to $\mathcal{D}_{t}$, only a small fraction of its predictions, characterized by the highest confidences and proximity to class centroids, may be reliable. As the model gradually learns the distribution of $\mathcal{D}_{t}$, it becomes capable of providing reliable predictions for an increasing number of pixels, as shown in Fig. \ref{figure:prepseudolabel}. Additionally, this mechanism eliminates the need for empirically setting thresholds or weights, avoiding the introduction of additional hyperparameters requiring manual tuning.

\subsection{Cross-domain overall loss function}
\label{sec:lossfunction}
In the actual land cover distribution, there may be a severe issue of class imbalance. In extreme cases, the most common category may cover areas that are hundreds of times larger than some other categories \cite{SSL}. To further alleviate this problem, we introduce class-balanced weights when constructing the loss function. Unlike previous class-balancing method computing weights based on the global distribution \cite{FBP}, we calculate class weights for each mini-batch, emphasizing small sparse labels in local regions. In the target branch, we can obtain the segmentation loss:

\begin{equation}
\mathcal{L}_{S}^{t}=\sum_{k=1}^{K}\frac{1}{\log{(\mu_{k}+1)}}\ell_{CE}(\bar{y}_{t}^{(k)},\bar{p}_{t}^{(k)}),
\end{equation}

where $\bar{p}_{t}^{(k)}=\{{p}_{t}^{(i,k)}|{i}\in\Omega_{l}\cup\Omega_{e}\}$, and $\ell_{CE}$ is the standard cross-entropy segmentation loss. $\mu_{k}$ is the proportion of $k$th class samples in the current mini-batch, denoted as

\begin{equation}
\mu_{k}=\frac{\sum_{i}{\mathbb{I}}(\bar{y}_{t}^{(i,k)}==1)}{\sum_{j}\sum_{q}{\mathbb{I}}(\bar{y}_{t}^{(j,q)}==1)}.
\end{equation}

Since $\bar{y}_{t}$ is a combination of true labels and pseudo-labels, this loss encompasses both supervised training and self-training.

For the unlabeled region $\Omega_{u}$ in ${x}_{t}$, although this area lacks explicit supervision and not all pixels within it will be assigned pseudo-labels, we still want to learn information from it to adapt to $\mathcal{D}_{t}$. Considering that the model is expected to directly generate accurate predictions in the late stages of domain adaptation, this implies the need to minimize the difference in prediction probabilities before and after rectification. Based on this thought, we define a rectification loss:

\begin{equation}
\mathcal{L}_{R}^{t}=\frac{1}{|\Omega_{u}|}\sum_{{i}\in\Omega_{u}}|{p}_{t}^{(i)}-{w}_{t}^{(i)}{p}_{t}^{(i)}|.
\end{equation}

where $|\Omega_{u}|$ denotes the number of pixels in the unlabeled area, this is adopted to control the losses at a similar magnitude.

Finally, the cross-domain overall loss can be calculated as

\begin{equation}
\mathcal{L}_{total}=\mathcal{L}^{s}+\mathcal{L}_{S}^{t}+\mathcal{L}_{R}^{t},
\end{equation}

where $\mathcal{L}^{s}$ is the cross-entropy segmentation loss for the source branch, similarly computed with class-balanced weighting.

\subsection{Dynamic prototype update}
\label{sec:update}
As training progresses, the model gradually becomes acquainted with the distribution $\mathcal{D}_{t}$, allowing it to generate more accurate deep features for ${x}_{t}$. Therefore, it is necessary to update the initial prototypes with the newly acquired features. To enhance computational efficiency, we update the prototypes with ${y}_{t}$ of the current mini-batch \cite{proda}. Particularly, in each iteration, the new prototype is estimated by moving the cluster centroid as 

\begin{equation}
\eta^{(k)}\gets\lambda\eta^{(k)}+(1-\lambda)\eta^{'(k)},
\end{equation}

where $\lambda$ is the momentum coefficient customarily set to 0.999, and $\eta^{'(k)}$ is calculated with the current training batch $\mathcal{X}_{t}^{'}\subset\mathcal{X}_{t}$ of the target branch. The updated prototypes are used for the rectification and expansion of pseudo-labels in the next iteration. During this process, both the prototypes and pseudo-labels can be iteratively refined.

\section{Experiments}
\subsection{Experimental setup}
\label{sec:setup}
\textbf{Data processing.} To transfer the models to satellite images of different resolutions, we match the Five-Billion-Pixels dataset with other satellite images by constructing a multi-scale $\mathcal{D}_{s}$. Depending on the spatial resolution of different data sources, image patches with different sizes are randomly cropped from GF-2, including $512\times 512$, $1024\times 1024$ (to match GF-1), and $1280\times 1280$ (to match ST-2) pixels, and then uniformly resized them to $512\times 512$ pixels. Since the $3$ m resolution of PS is achieved by resampling the raw data, its effective spatial resolution is $3.7$-$4.1$ m. We use the original image resolution of GF-2 for adaptation. The total number of patches in $\mathcal{D}_{s}$ is $3.2\times 10^{4}$, with a ratio of $2:1:1$ for the three sizes, which are first used to pre-train the backbone and then serve as inputs for the source branch.

The satellite images in C-megacities and G-cities are cropped into patches of size $512\times 512$ pixels. Specifically, PS images are resized to $3/4$ of the original image resolution before cropping, roughly equivalent to restoring their spatial resolution to $4$ m. Due to the different category systems in C-megacities and G-cities, we train segmentation models separately for each of them.

As the number of patches in $\mathcal{D}_{s}$ far exceeds that of $\mathcal{D}_{t}$, for each epoch of training, an equal number of patches are randomly selected from $\mathcal{D}_{s}$ as in $\mathcal{D}_{t}$. This results in different subsets of $\mathcal{D}_{s}$ for each epoch, allowing the model to learn diverse information.

\textbf{Category alignment.} To transfer knowledge from Five-Billion-Pixels to G-cities, which have different category systems, we roughly match the classes from Five-Billion-Pixels to those of G-cities. Specifically, for Five-Billion-Pixels, \textit{dry cropland} and \textit{irrigated field} are merged into \textit{cultivated land}; \textit{garden land} and \textit{arbor forest} are merged into \textit{forest}; \textit{lake} and \textit{pond} are merged into \textit{water body}; the \textit{park} category is removed, leaving 20 classes in Five-Billion-Pixels. The remaining differently named classes and their correspondence to G-cities are as follows: \textit{industrial area} - \textit{industrial/commercial}, \textit{urban residential} - \textit{high/mid residential}, \textit{rural residential} - \textit{low residential}, \textit{stadium} - \textit{public unit}, \textit{square} - \textit{public square}, \textit{river} - \textit{water course}. Categories present in G-cities but absent in Five-Billion-Pixels are \textit{paved area}, \textit{wetland}, \textit{mineral site} and \textit{construction site}, meaning that these classes are defined in the model's output layer, but there are no corresponding samples in the source branch. It is worth noting that even with rough correspondence, significant variations in the characteristics of corresponding classes exist between different domains due to disparities in geographical distribution and cultural background, necessitating the use of domain adaptation algorithms to bridge them.

\textbf{Comparison methods.} Since there is currently no research specifically addressing WSDA in the remote sensing community, we modify state-of-the-art WSSS methods into the form of WSDA by augmenting them with information from $\mathcal{D}_{s}$. The comparison includes DBFNet \cite{wsssDBFNet}, CRGNet \cite{wsssCRGNet}, and FESTA \cite{wsssFESTA}. The specific operation is to randomly select the same number of patches from $\mathcal{D}_{s}$ as in $\mathcal{D}_{t}$ for each epoch and merge the two domains as training inputs. The loss functions of the comparative methods are applied to $\mathcal{D}_{t}$, with the standard segmentation loss of $\mathcal{D}_{s}$ added on this basis.

DeepLabv3+ \cite{deeplab} and U-Net \cite{unet} pre-trained on $\mathcal{D}_{s}$ are used as backbones for CRGNet, FESTA, and PRE, respectively. Given the fixed design of the encoder and decoder in DBFNet, which cannot be replaced, we use ResNet-101 \cite{resnet} pre-trained on $\mathcal{D}_{s}$ as its backbone.

All training hyperparameters are set consistently. The batch size for both source and target branches is 12 (total 24). The initial learning rate is 0.005, the number of epochs is 100, momentum is 0.9, weight decay is $1\times 10^{-4}$, and the poly learning strategy with a power of 0.9 is used to adjust the learning rate during epochs.

\textbf{Evaluation metrics.} We assess the experimental results using overall accuracy (OA), mean F1 score (mF1), mean intersection over union (mIoU), and producer's accuracy (PA). mF1 is the class-wise average of F1-scores. mIoU is the class-wise mean of intersection over union (IoU), where IoU is calculated by dividing the intersection of prediction and ground truth by their union. mF1 and mIoU describe the model's ability to minimize overestimation and underestimation for each class. PA represents the model's performance in reducing underestimation \cite{producer}.

\begin{table*}[htb!]
\centering
\caption{Comparative results on C-megacities and G-cities. ``RN101'', ``DLv3+'' and ``UN'' respectively represent ResNet-101, DeepLabv3+, and U-Net. Accuracy values are expressed as percentage (\%). The best score for each column is highlighted.}
\vspace{3mm}
\resizebox{0.9\textwidth}{!}{
\begin{tabular}{lllllllllllllll} 
\hline
\multirow{3}{*}{Method} & \multicolumn{6}{l}{\textbf{C-megacities}}                                                           &  & \multicolumn{6}{l}{\textbf{G-cities}}                                                               \\
                        & \multicolumn{3}{l}{Sparse Label}                 & \multicolumn{3}{l}{Dense Label}                  &  & \multicolumn{3}{l}{Sparse Label}                 & \multicolumn{3}{l}{Dense Label}                  \\ \cline{2-7} \cline{9-14} 
                        & OA             & mF1            & mIoU           & OA             & mF1            & mIoU           &  & OA             & mF1            & mIoU           & OA             & mF1            & mIoU           \\ \hline
DBFNet-RN101           & 79.12          & 52.01          & 40.05          & 67.56          & 39.46          & 28.99          &  & 74.96          & 40.22          & 31.37          & 69.72          & 34.09          & 24.60          \\
CRGNet-DLv3+           & 89.62          & 67.82          & 57.70          & 82.57          & 58.28          & 48.29          &  & 86.65          & 55.89          & 46.79          & 80.53          & 54.35          & 42.55          \\
CRGNet-UN              & 90.96          & 67.46          & 56.37          & 83.17          & 58.68          & 48.34          &  & 86.82          & 57.23          & 46.75          & 79.94          & 58.28          & 45.35          \\
FESTA-DLv3+            & 91.60          & 71.60          & 61.54          & 83.19          & 59.14          & 49.07          &  & 84.70          & 54.01          & 44.40          & 80.12          & 53.41          & 41.70          \\
FESTA-UN               & 91.81          & 69.42          & 58.81          & 83.27          & 60.08          & 49.10          &  & 87.51          & 59.77          & 49.08          & 80.78          & 59.70          & 47.42          \\
PRE-DLv3+ (ours)    & \textbf{92.88}	& \textbf{72.72} & \textbf{63.28} & 83.32          & \textbf{60.47}	& \textbf{50.47} &  & \textbf{88.17} & \textbf{64.03} & \textbf{53.94} & \textbf{83.24} & \textbf{64.74} & \textbf{52.33} \\
PRE-UN (ours)       & 92.56          & 69.96          & 59.45          & \textbf{83.70} & 56.72          & 46.06          &  & 87.36          & 62.39          & 51.52          & 82.68          & 63.14          & 51.06          \\ 
\hline
\end{tabular}}
\label{table:resultcomparison}
\end{table*}

\subsection{Comparative experiments}
Table \ref{table:resultcomparison} presents the comparative results with different methods. Overall, our PRE outperforms all comparative methods, particularly demonstrating significant improvements for G-cities. Specifically, on the sparse test labels of G-cities, PRE surpasses the state-of-the-art by $4.86\%$ in mIoU, while on the dense test labels, PRE outperforms the state-of-the-art by $4.91\%$ in mIoU. These improvements are attributed to the domain gap between $\mathcal{D}_{s}$ and $\mathcal{D}_{t}$. G-cities and Five-Billion-Pixels are situated on different continents with some differences in category systems. In this scenario where there is a significant gap between $\mathcal{D}_{s}$ and $\mathcal{D}_{t}$, our method benefits greatly from capturing the global feature distribution of $\mathcal{D}_{t}$ for model transfer. However, despite the smaller domain difference between C-megacities and Five-Billion-Pixels, improvements of over $1\%$ in mIoU are still achieved on both the sparse and dense test labels of C-megacities. This underscores the effectiveness of our approach in bridging domain gaps and enhancing semantic segmentation performance across diverse feature distributions.

Among the comparative methods, DBFNet \cite{wsssDBFNet} establishes constraint relationships for the edges of objects, which is effective for very-high-resolution ($<1$ m) imagery. However, for land cover classification at the meter level, the boundaries of objects are often not very clear, and differences in spectral response between different domains can greatly affect the performance of different object boundaries. CRGNet \cite{wsssCRGNet} expands pseudo-labels based on pixel neighborhoods and constrains the consistency of results from different training branches. This strategy may not be ideal for large-scale data classification because in large areas, objects belonging to the same class may not be located in adjacent regions. FESTA \cite{wsssFESTA} models constraint relationships for pixels within the same image, which cannot capture the feature distribution information of the entire domain and may have difficulty dealing with intra-domain and inter-domain differences.

\begin{figure*}[htb!]
\centering
\includegraphics[width=0.9\textwidth]
{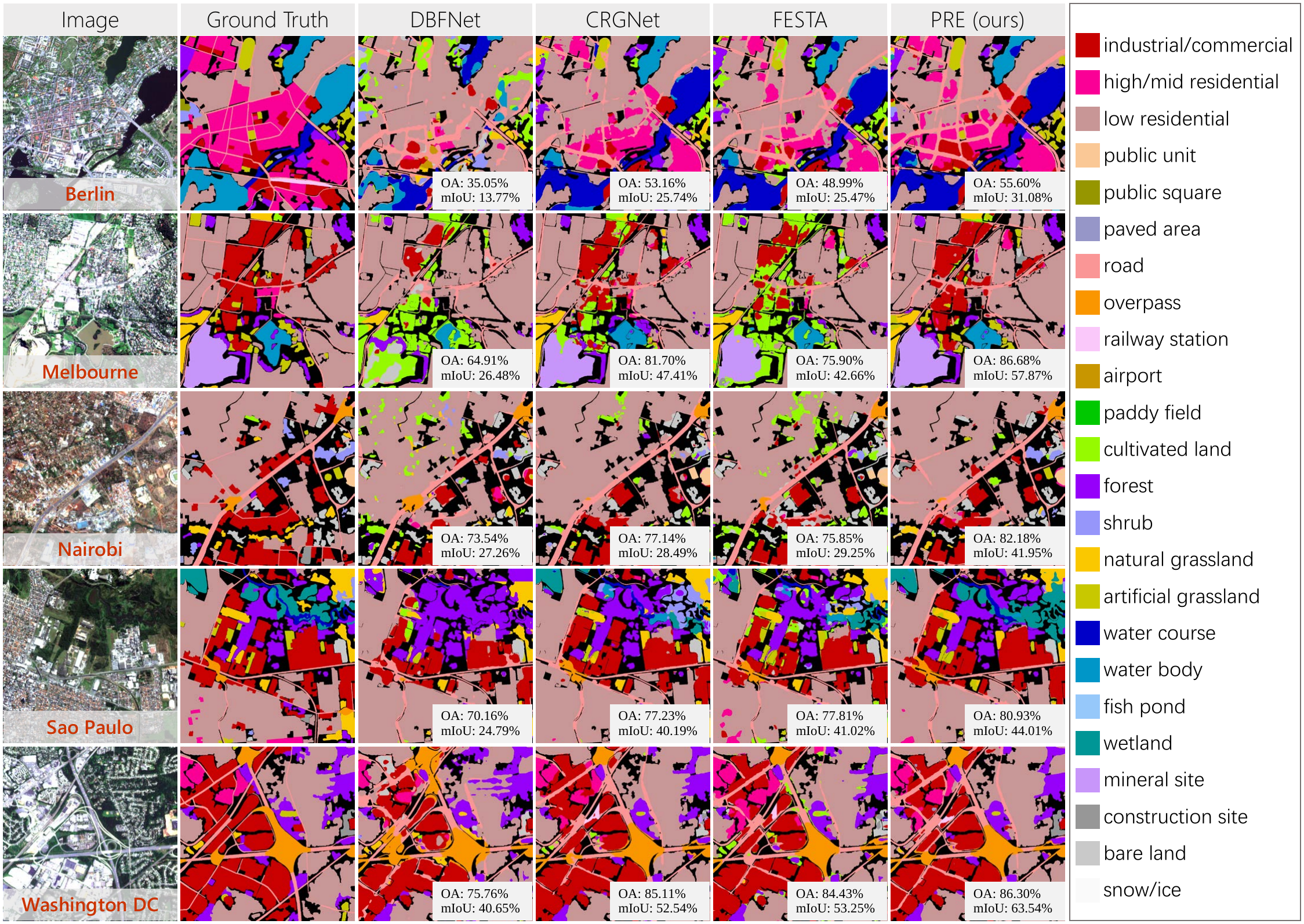}
\caption{The detailed performance of different methods on the dense test labels in G-cities. The backbone is DeepLabv3+.}
\label{figure:resultg}
\end{figure*}

It is noted that different backbones also contribute to performance differences. DeepLabv3+ outperforms on both test sets, as it adopts a deep network structure that can capture more abstract contextual features of ground objects. In situations with a domain gap, these features are beneficial for domain transfer, resulting in better overall and local boundary adaptation. In general, our method demonstrates stable performance across different backbones, with improvements observed in both DeepLabv3+ and U-Net.

Details of the results for G-cities can be observed in Fig. \ref{figure:resultg}. Our method significantly enhances the results for residential areas in Berlin, industrial areas and artificial grasslands in Melbourne, industrial areas and shrubs in Nairobi, wetlands and natural grasslands in Sao Paulo, and forests and roads in Washington DC. Moreover, compared to the comparative methods, our PRE notably reduces segmentation noise at the edges of different land cover types. This indicates the strong capability of PRE to transfer information from $\mathcal{D}_{s}$, especially regarding ground object boundaries, to $\mathcal{D}_{t}$. Additionally, PRE can accurately capture the global feature distribution of $\mathcal{D}_{t}$ based solely on its sparse labels.

The PA values for different categories on the sparse test labels of G-cities are shown in Table \ref{table:resultgcategories}, with ResNet-101 and DeepLabv3+ as the backbones. It can be observed that out of the 23 categories present in the test data, our PRE achieves the best accuracy in 16 categories. Moreover, PRE shows over $10\%$ improvement in categories such as paved area, paddy field, shrub, artificial grassland, mineral site, and construction site. It is noteworthy that paved area, wetland, mineral site, and construction site are categories not defined in Five-Billion-Pixels, highlighting the superiority of PRE in cross-category domain adaptation. In some categories, our method fails to identify any correct pixels or performs significantly worse than comparative methods, such as public unit, public square, and fish pond. This is because these categories have very few pixels annotated in the sparse labels, as shown in Table \ref{table:percentage}. In the absence of sufficient samples, PRE struggles to construct accurate class prototypes, thereby hindering pseudo-label refinement and domain adaptation.

\begin{table*}[htb!]
\centering
\caption{The PA values for different categories on the sparse test labels of G-cities under different methods. ``RN101'' and ``DLv3+'' respectively represent ResNet-101 and DeepLabv3+. ``*'' indicates classes not defined in $\mathcal{D}_{s}$, while ``$^+$'' indicates classes with a pixel proportion of less than $1\%$ in Table \ref{table:percentage}. ``-'' indicates that there are no samples of this class in the test data. Accuracy values are expressed as percentage (\%).}
\vspace{3mm}
\resizebox{0.9\textwidth}{!}{
\begin{tabular}{lllllllllllll}
\hline
Method        & Inco        & Hres        & Lres        & Puni$^+$    & Psqu$^+$    & Pave*$^+$    & Road        & Over$^+$          
              & Rail$^+$    & Airp        & Padd$^+$    & Cult           \\ \hline
DBFNet-RN101 & 73.83          & 13.55          & 89.93          & 1.54           & 0              & 0              & 82.16          & 59.96          & 0.41           & 51.74          & 0              & 79.31          \\
CRGNet-DLv3+ & 87.61          & 63.92          & \textbf{98.38} & 63.52          & 0              & 6.15           & 91.52          & 82.18          & 12.11          & 90.27          & 1.44           & 87.12          \\
FESTA-DLv3+  & 86.35          & 58.61          & 96.83          & \textbf{63.72} & \textbf{0.05}  & 0.31           & 88.66          & 61.52          & 14.90          & 79.58          & 0.37           & \textbf{93.58} \\
PRE-DLv3+ (ours)  & \textbf{92.28} & \textbf{67.10} & 98.20          & 50.00          & 0              & \textbf{44.68} & \textbf{95.11} & \textbf{83.82} & \textbf{19.49} & \textbf{99.61} & \textbf{20.97} & 82.68          \\ \hline
              & Fore        & Shru        & Ngra        & Agra        & Wcou        & Wbod         & Fish$^+$    & Wetl*          
              & Mine*       & Cons*$^+$   & Bare        & Snic           \\ \hline
DBFNet-RN101 & 80.00          & 41.12          & 65.47          & 27.09          & 16.76          & 91.34          & 0              & 0.90           & 23.68          & 0.18           & 75.25          & -              \\
CRGNet-DLv3+ & 93.54          & 70.31          & 74.34          & 55.52          & 29.78          & \textbf{94.01} & \textbf{0.49}  & 7.72           & 56.82          & 3.92           & \textbf{87.95} & -              \\
FESTA-DLv3+  & 90.31          & 57.25          & 58.31          & 56.16          & 27.83          & 85.51          & 0              & 4.80           & 67.58          & 7.57           & 85.98          & -              \\
PRE-DLv3+ (ours)  & \textbf{96.79} & \textbf{80.65} & \textbf{77.48} & \textbf{79.03} & \textbf{34.28} & 87.64          & 0              & \textbf{15.43} & \textbf{92.49} & \textbf{36.34} & 87.48          & -              \\ 
\hline
\end{tabular}}
\label{table:resultgcategories}
\end{table*}

Another notable issue is the significant performance disparity among different categories, regardless of the method type. For instance, all methods perform poorly in categories like public square, paved area, railway station, paddy field, wetland, and construction site, which is attributed to two factors. First, these categories account for only a small percentage in weak labels, causing the model to bias towards dominant categories. Second, they are inherently more prone to confusion with other categories. For example, railway station includes multiple tracks and structures resembling stadiums, making it susceptible to misclassification as other categories. Similarly, wetland, being a category not defined in Five-Billion-Pixels, is easily confused with natural grassland, despite anchoring its feature distribution with sparse annotations from $\mathcal{D}_{t}$. However, due to our consideration of class imbalance in the design of PRE, the results of these challenging categories generally show notable improvements, confirming the practicality of our strategy.

\begin{table*}[htb!]
\centering
\caption{Ablation study regarding data. ``S Only'' denotes training the model using $\mathcal{D}_{s}$ only; ``T Only'' indicates training the model using $\mathcal{D}_{t}$ only; ``T PRE'' involves using a model pre-trained on $\mathcal{D}_{s}$ to perform PRE, and $\mathcal{D}_{s}$ participating in pre-training but not in the training of PRE; ``S+T PRE'' represents our method. Accuracy values are expressed as percentage (\%).}
\vspace{3mm}
\resizebox{0.9\textwidth}{!}{
\begin{tabular}{llllllllllllll}
\hline
\multirow{3}{*}{Data} & \multicolumn{6}{l}{\textbf{C-megacities}}                                                           &  & \multicolumn{6}{l}{\textbf{G-cities}}                                                               \\
                      & \multicolumn{3}{l}{Sparse Label}                 & \multicolumn{3}{l}{Dense Label}                  &  & \multicolumn{3}{l}{Sparse Label}                 & \multicolumn{3}{l}{Dense Label}                  \\ \cline{2-7} \cline{9-14} 
                      & OA             & mF1            & mIoU           & OA             & mF1            & mIoU           &  & OA             & mF1            & mIoU           & OA             & mF1            & mIoU           \\ \hline
S Only-DLv3+          & 85.90          & 61.89          & 51.27          & 80.45          & 52.19          & 42.30          &  & 73.64          & 39.60          & 31.21          & 71.01          & 35.70          & 26.29          \\
S Only-UN             & 84.69          & 59.25          & 47.21          & 80.38          & 51.66          & 42.27          &  & 63.36          & 32.59          & 24.06          & 68.66          & 35.72          & 25.91          \\ \hline
T Only-DLv3+          & 71.04          & 42.55          & 30.95          & 54.01          & 25.36          & 17.02          &  & 76.66          & 40.13          & 30.90          & 72.21          & 41.70          & 30.98          \\
T Only-UN             & 85.31          & 58.71          & 47.16          & 70.42          & 39.04          & 28.60          &  & 82.04          & 51.35          & 40.56          & 77.27          & 52.25          & 39.55          \\ \hline
T PRE-DLv3+           & 92.59          & 71.11          & 61.03          & 76.25          & 53.32          & 42.75          &  & 85.71          & 59.99          & 49.23          & 81.49          & 60.33          & 48.39          \\
T PRE-UN              & 92.30          & 69.53          & 59.71          & 79.71          & 49.83          & 38.54          &  & 86.51          & 57.32          & 46.25          & 82.16          & 60.37          & 48.12          \\ \hline
S+T PRE-DLv3+         & \textbf{92.88} & \textbf{72.72} & \textbf{63.28} & 83.32          & \textbf{60.47} & \textbf{50.47}          &  & \textbf{88.17} & \textbf{64.03} & \textbf{53.94} & \textbf{83.24} & \textbf{64.74} & \textbf{52.33} \\
S+T PRE-UN            & 92.56          & 69.96          & 59.45          & \textbf{83.70} & 56.72          & 46.06          &  & 87.36          & 62.39          & 51.52          & 82.68          & 63.14          & 51.06          \\ 
\hline
\end{tabular}}
\label{table:resultablation}
\end{table*}

\subsection{Ablation study}
To further analyze the mechanism of PRE, we perform ablation studies regarding both data and algorithm components. Regarding the data, we conduct experiments using only $\mathcal{D}_{s}$, only $\mathcal{D}_{t}$, and a combination of both domains, with the results listed in Table \ref{table:resultablation}. Specifically, ``S Only'' denotes training the model using Five-Billion-Pixels and directly applying it to segment C-megacities and G-cities; ``T Only'' indicates training the model using only the sparse annotations of $\mathcal{D}_{t}$ and generating results; ``T PRE'' involves using a model pre-trained on $\mathcal{D}_{s}$ to perform PRE, retaining only the target branch, meaning $\mathcal{D}_{s}$ participates in pre-training but not in the training of PRE; ``S+T PRE'' represents our method. Similarly, DeepLabv3+ and U-Net are used as the backbones. It can be observed that combining both domains leads to a significant improvement in all evaluation metrics. This suggests that combining dense labels from $\mathcal{D}_{s}$ and sparse labels from $\mathcal{D}_{t}$ is necessary for reducing the annotation cost of new data while ensuring the accuracy of classification results. For C-megacities, the performance of ``S Only'' is better than ``T Only'' because at this point, the distributions of $\mathcal{D}_{s}$ and $\mathcal{D}_{t}$ are closer, and the complete annotations of $\mathcal{D}_{s}$ better guide the model learning compared to the sparse annotations $\mathcal{D}_{t}$. However, for G-cities, ``T Only'' outperforms ``S Only'' because although the sparse annotations are incomplete, they represent the feature distribution of $\mathcal{D}_{t}$, indicating significant domain differences between Five-Billion-Pixels and G-cities. ``T PRE'' outperforms both ``S Only'' and ``T Only'' on all datasets, indicating that using a model pre-trained on the source domain can guide the generation of prototypes in $\mathcal{D}_{t}$ to some extent.

\begin{figure*}[htb!]
\centering
\includegraphics[width=0.9\textwidth]
{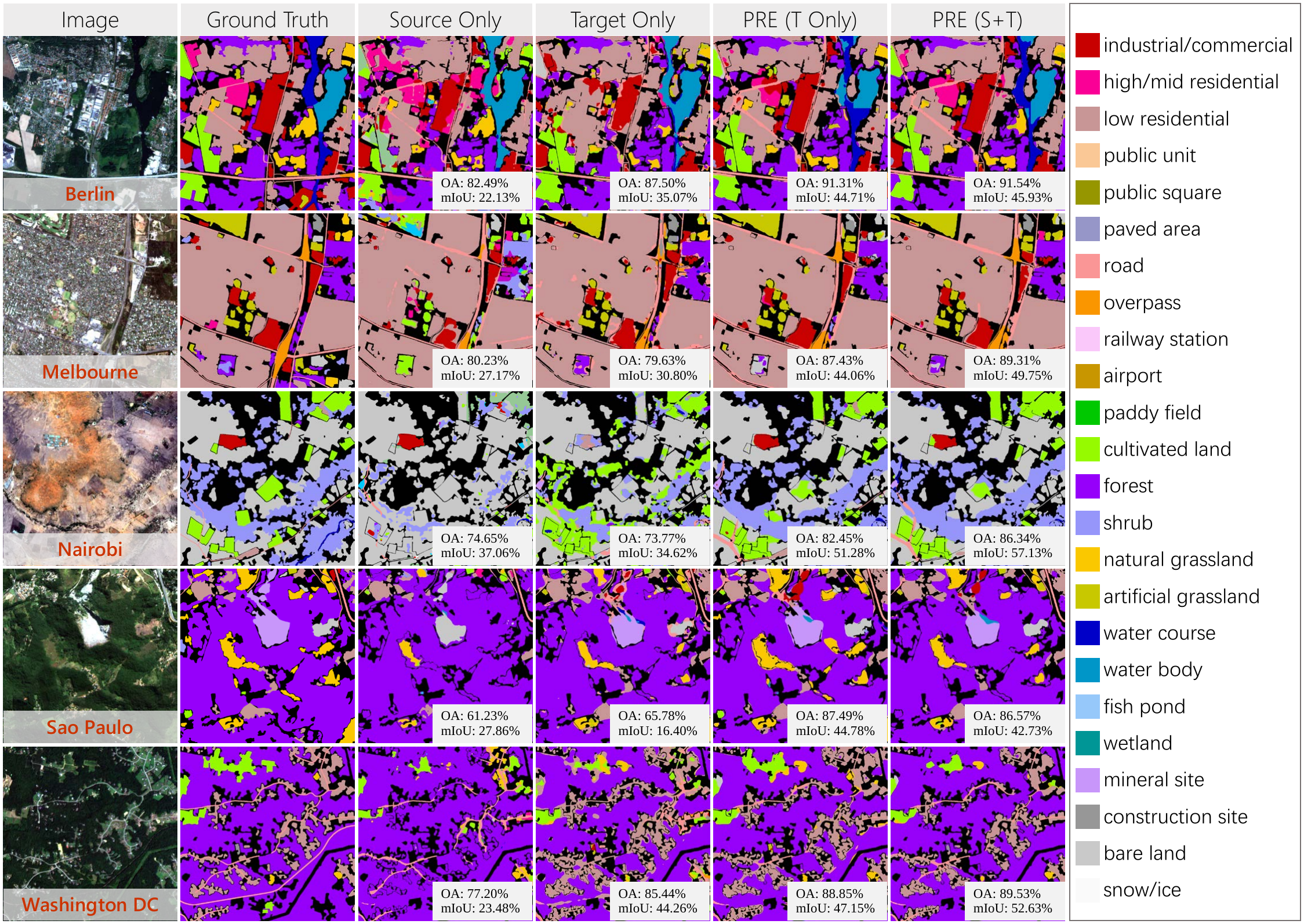}
\caption{The detailed performance of ablation study regarding data in G-cities. The backbone is DeepLabv3+. ``Source Only'' denotes training the model using $\mathcal{D}_{s}$ only; ``Target Only'' indicates training the model using $\mathcal{D}_{t}$ only; ``PRE (T Only)'' involves using a model pre-trained on $\mathcal{D}_{s}$ to perform PRE on $\mathcal{D}_{t}$ only; ``PRE (S+T)'' represents our method.}
\label{figure:resultablation}
\end{figure*}

The detailed results of the data ablation study on G-cities are illustrated in Fig. \ref{figure:resultablation}. It can be observed that the ``S Only'' results exhibit some notable misclassifications, such as mislabeling artificial grassland as cultivated land in Melbourne and mislabeling shrub as bare land in Nairobi. On the other hand, ``T Only'' tends to generate segmentation noise, as seen in Berlin, where the industrial area's results are fragmented and disorderly. ``T PRE'' shows significant improvements compared to the former two, indicating that $\mathcal{D}_{s}$ can effectively guide the generation of prototypes and reduce domain shift. ``S+T PRE'' demonstrates the most accurate land cover classification and edge segmentation results, highlighting the necessity of $\mathcal{D}_{s}$ for preserving land boundaries and $\mathcal{D}_{t}$ for capturing land features. These results indicate that our method PRE, as a form of WSDA, can effectively mitigate the reliance on high-quality labels for accurate land cover classification.

\begin{table*}[htb!]
\centering
\caption{Ablation study for algorithm components on G-cities, with DeepLabv3+ as the backbone. $\mathcal{L}^{s}$: segmentation loss of the source branch; $\mathcal{L}_{S}^{t}$: expansion segmentation loss of the target branch; $\mathcal{L}_{R}^{t}$: rectification loss of the target branch; ``CB'': class balance weighting. Accuracy values are expressed as percentage (\%).}
\vspace{3mm}
\resizebox{0.6\textwidth}{!}{
\begin{tabular}{llllllll}
\hline
\multirow{2}{*}{Components} & \multicolumn{3}{l}{Sparse Label} &  & \multicolumn{3}{l}{Dense Label} \\ \cline{2-4} \cline{6-8} 
                            & OA        & mF1       & mIoU     &  & OA        & mF1      & mIoU     \\ \hline
$\mathcal{L}^{s}$                                                                                & 73.64     & 39.60     & 31.21    &  & 71.01     & 35.70    & 26.29    \\
$\mathcal{L}^{s}$ + $\mathcal{L}_{S}^{t}$                                            & 87.03     & 60.26     & 50.26    &  & 82.60     & 61.57    & 49.52    \\
$\mathcal{L}^{s}$ + $\mathcal{L}_{S}^{t}$   + CB                                     & 87.08     & 63.03     & 52.73    &  & 82.44     & 62.71    & 50.31    \\
$\mathcal{L}^{s}$ + $\mathcal{L}_{S}^{t}$   + CB + $\mathcal{L}_{R}^{t}$ & \textbf{88.17}     & \textbf{64.03}     & \textbf{53.94}    &  & \textbf{83.24}     & \textbf{64.74}    & \textbf{52.33}    \\ 
\hline
\end{tabular}}
\label{table:resultgcomponents}
\end{table*}

Regarding the algorithm components, we conduct ablation study on G-cities using different combinations to observe their impact on the results, as shown in Table \ref{table:resultgcomponents}. Here, $\mathcal{L}^{s}$ represents the segmentation loss of the source branch, $\mathcal{L}_{S}^{t}$ denotes the expansion segmentation loss of the target branch, ``CB'' refers to class balance weighting, and $\mathcal{L}_{R}^{t}$ represents the rectification loss of the target branch. It can be observed that incorporating each component leads to performance improvement. $\mathcal{L}_{S}^{t}$ enables the model to learn the distribution of $\mathcal{D}_{t}$ and adapt to it. ``CB'' helps alleviate the challenging issue of class imbalance in real-world land cover mapping. $\mathcal{L}_{R}^{t}$ further aligns the feature distribution of $\mathcal{D}_{t}$ with the distribution of class prototypes. Our approach performs best when all the components are utilized simultaneously, indicating that these components complement each other and are crucial for enhancing performance.

\begin{figure*}[htb!]
\centering
\includegraphics[width=0.9\textwidth]
{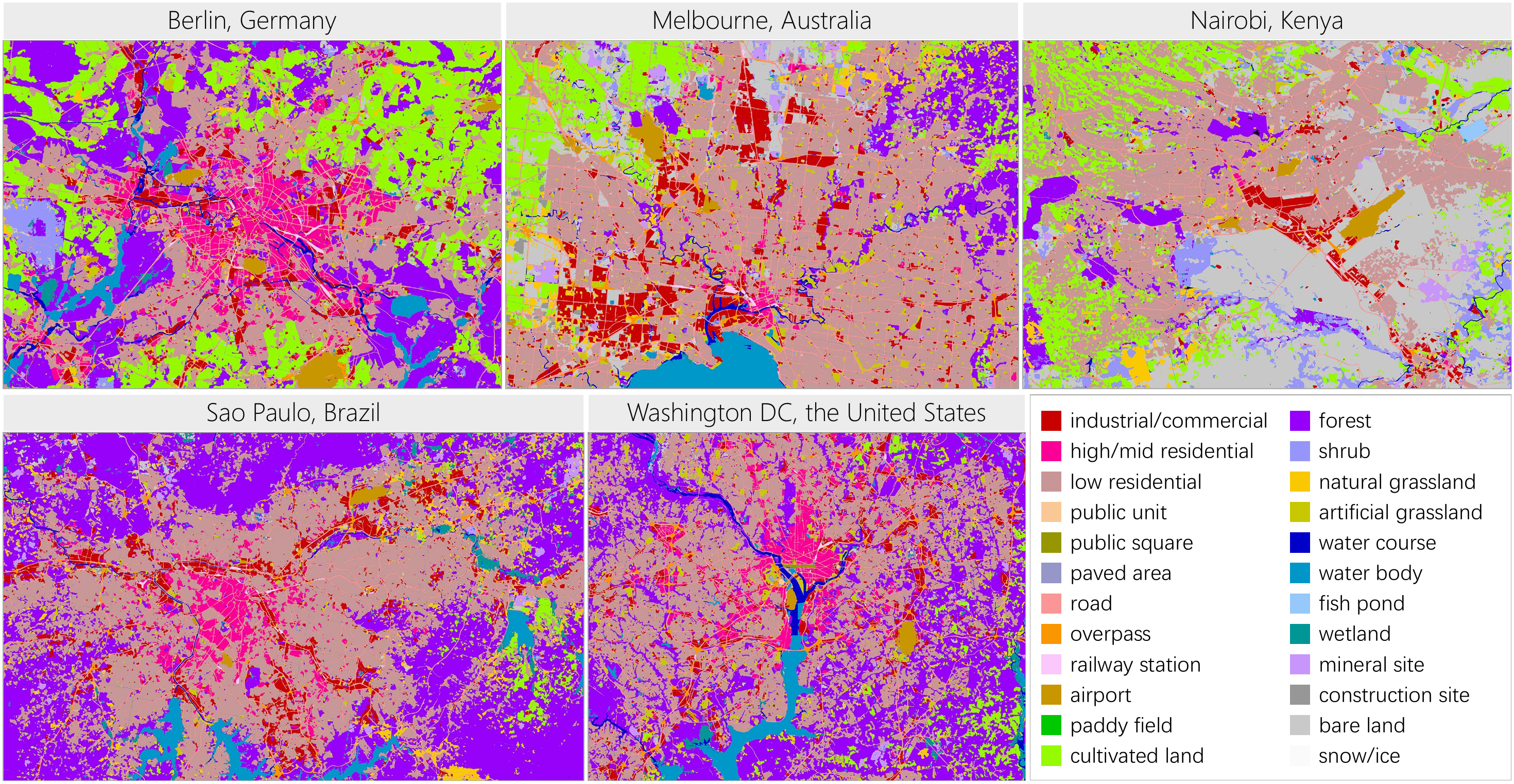}
\caption{Land cover mapping results for G-cities.}
\label{figure:mappingg}
\end{figure*}

\subsection{Land cover mapping}
The land cover mapping results of G-cities are illustrated in Fig. \ref{figure:mappingg}. These results are mosaicked from 101 PS images, with 30 images having weak annotations used for model training, 10 images having test annotations but not involved in training, and the remaining 71 images having no annotations at all. To ensure the completeness of visual presentation, these 30 training images are not removed during mapping. Therefore, these results are presented for illustrative purposes only and are not quantitatively evaluated. It can be observed that our results distinguish industrial/commercial areas, high/mid-rise residential areas in city centers, and low-rise residential areas in the surrounding regions. The transportation networks and river systems of each city are correctly identified. Agricultural areas and forests on the outskirts of cities are clearly classified. These results confirm the domain adaptation capability of our method across sensors, categories, and continents, with the potential for large-scale land cover mapping in different regions globally.

\section{Discussion}
\subsection{Potential for specific land type detection}
\begin{figure*}[htb!]
\centering
\includegraphics[width=0.9\textwidth]
{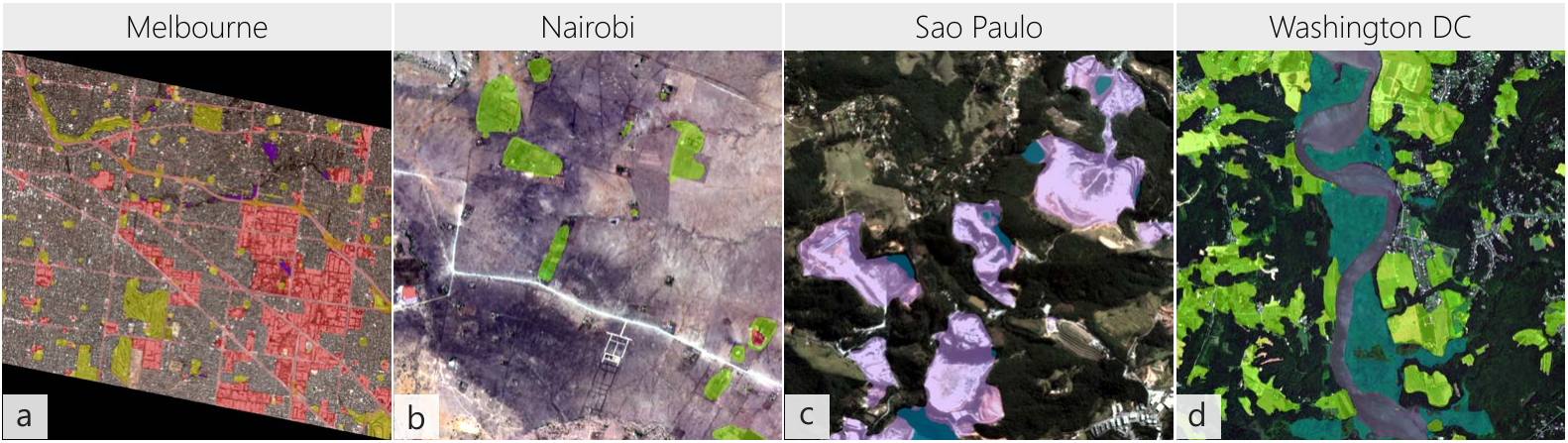}
\caption{The detailed land cover mapping results depict specific categories relevant to the natural environment, economic activities, and urban quality of life, with other categories hidden for visual clarity. (a) Urban industrial zones and urban green spaces in Melbourne; (b) Smallholder agriculture in Nairobi; (c) Mineral sites within forested areas in Sao Paulo; (d) Suburban croplands and wetlands in Washington DC.}
\label{figure:mappingdetail}
\end{figure*}

Accurate measurements of environmental conditions and economic livelihoods have significant implications for both research and policy making. And this is particularly crucial in many developing countries, where such data are often lacking \cite{ethical,importance2}. Given the challenges of traditional data collection methods, satellite remote sensing imagery has emerged as a valuable solution for environmental and economic assessments. In this discussion, we demonstrate the potential of our work in acquiring specific categories relevant to the environment and economy. Fig. \ref{figure:mappingdetail} depict the segmentation results obtained using PRE: (a) Urban industrial zones and urban green spaces in Melbourne, Australia; (b) Smallholder agriculture in Nairobi, Kenya; (c) Mineral sites within forested areas in Sao Paulo, Brazil; (d) Suburban croplands and wetlands in Washington DC, the United States.

Smallholder agriculture is pivotal for securing food supplies and reducing poverty, especially in rural regions of developing nations. Mineral sites are indispensable for a range of industrial activities such as manufacturing, construction, and energy generation, serving as key resources for economic progress. Wetlands are an essential part of ecological systems, promoting biodiversity and maintaining delicate ecological equilibrium.

It can be observed that our approach accurately detects these categories in cities located on different continents, despite the presence of interference from other categories within a complex category system. In future work, WSDA-based detection methods tailored to specific categories can be developed by treating other categories as background and performing binary segmentation. This has the potential to provide more accurate and comprehensive information to assist in environmental monitoring, natural resource management, and other fields.

\subsection{Potential for urban landscape analysis}
From our land cover mapping results in Fig. \ref{figure:mappingg}, we can observe the landscape layout of different cities.

Berlin presents a complex European old-town style, with extensive areas of compact high/mid-rise zones, fewer and scattered industrial areas, and satellite suburbs with evenly distributed farmlands surrounding the city. Melbourne exhibits a regular urban structure, characterized by straight roads and high/mid-rise buildings mainly concentrated in the port area. The city boasts a well-developed industrial sector, with industrial zones predominantly located in the southern part, alongside numerous urban green spaces. Nairobi lacks high/mid-rise areas, with industrial zones concentrated in the central part of the city. The suburbs feature a mix of low-rise residential areas and farmlands, arranged in a chaotic manner. Sao Paulo's high/mid-rise zones are centralized in the city center, while industrial areas extend along the main roads on both sides, with the city surrounded by forests and abundant wetlands in the northeast direction. Washington DC's high/mid-rise buildings exhibit a multi-center clustering pattern, with fewer industrial areas situated further away from the city center along the main roads, and urban green spaces are scattered throughout the city. The city is surrounded by forests, with sporadic small farmlands scattered within the forests.

These land cover mapping results demonstrate our method's ability to classify urban morphology accurately, capturing different types of urban landscape layouts, and providing essential information for urban development decision-making.

\begin{figure*}[htb!]
\centering
\includegraphics[width=0.9\textwidth]
{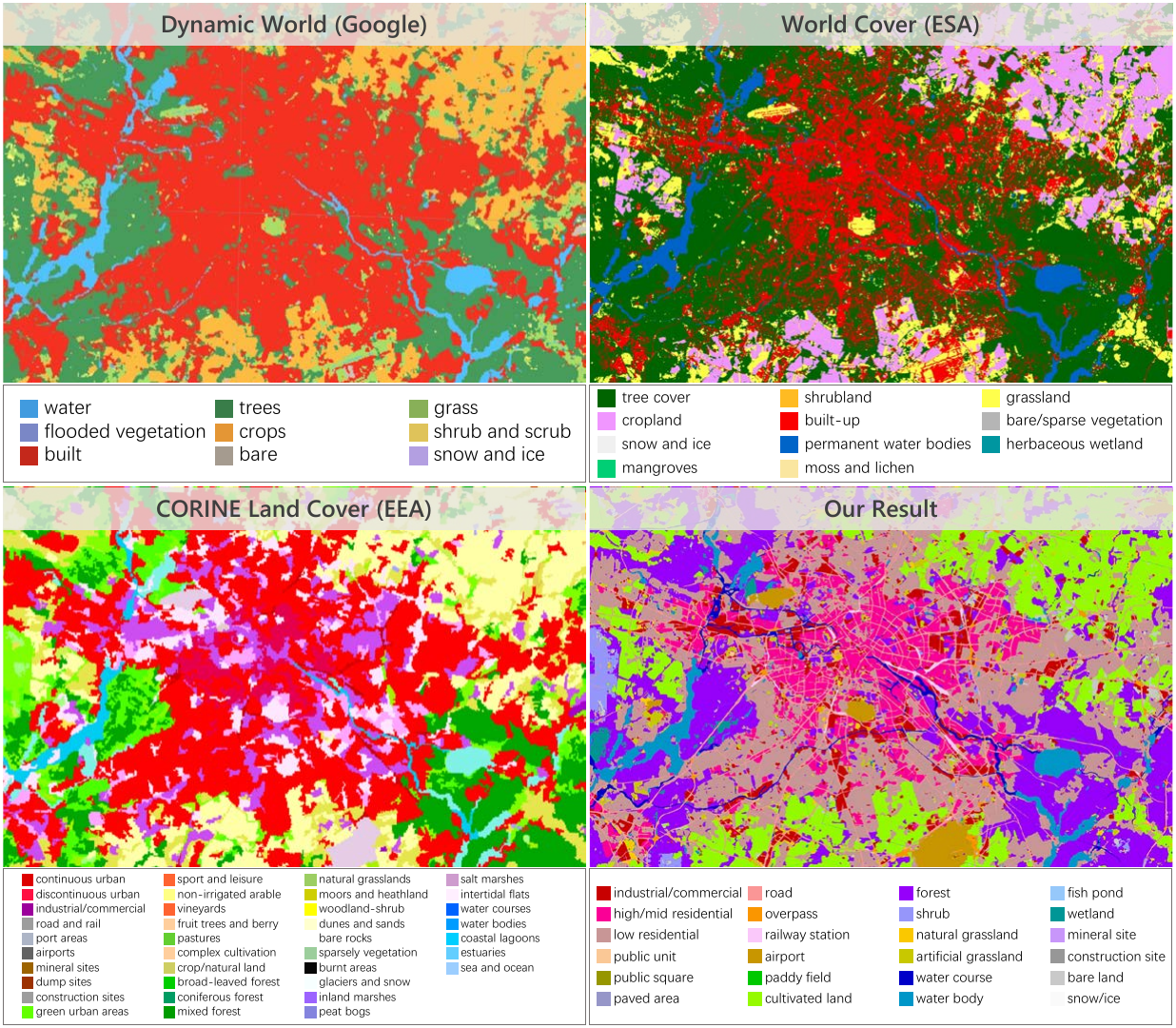}
\caption{Visual comparison with different land cover mapping projects in Berlin, including Google's Dynamic World, ESA's World Cover, and EEA's CORINE Land Cover.}
\label{figure:mappingberlin}
\end{figure*}

\subsection{Potential for large-scale land cover mapping}
We utilize weakly labeled satellite imagery to achieve fully automated classification, opening up new possibilities for real-time fine-grained land cover classification. Furthermore, our method has been demonstrated to be applicable across different sensors, categories, and continents, presenting new opportunities for global land cover mapping. 

We compare our results in Berlin with Google's Dynamic World \cite{googleLC}, ESA's World Cover \cite{esaLC}, and EEA's CORINE Land Cover \cite{corineLC}. Due to differences in data sources and acquisition times, we can only provide a rough visual comparison. As shown in Fig. \ref{figure:mappingberlin}, our results exhibit relatively consistent built-up areas, croplands, forests, and water bodies. Additionally, our results further differentiate land cover into more detailed categories, showcasing clear transportation networks, river systems, and various urban morphology and functional zones. Although we only study 10 cities in this work, our method can easily be extended to other cities, towns, and villages globally. Additionally, since our approach does not require precise full labels, it can easily leverage other ground landscape information currently available. For example, future work could involve obtaining local weak annotations (even with noise) using OpenStreetMap, Google Street View \cite{gsv}, or World Cover in conjunction with the fully labeled source domain to train transferable models.

One limitation is that we use PS satellite imagery for G-cities, while Dynamic World and World Cover utilize Sentinel satellite data, and CORINE Land Cover utilizes Landsat TM data. The higher resolution of PS imagery facilitates sparse labeling annotations and the study of WSDA, while the freely available Sentinel satellite data is more conducive to the development of large-scale land cover mapping research \cite{st2Mapping1,st2Mapping2,st2Mapping3}. In future studies, we hope to use high-resolution data for sparse labeling and transfer information to freely available ST-2 imagery, integrating WSDA and UDA into a unified framework, further promoting research in global fine-grained land cover mapping.

\section{Conclusion}
Leveraging remote sensing data for global fine-grained land cover mapping is a highly demanding task, as obtaining high-quality labeled data is extremely difficult, making research in this area challenging. In this paper, we present a WSDA approach, PRE, along with a set of weakly annotated datasets, C-megacities and G-cities, for large-scale high categorical resolution land cover mapping. By combining fully labeled source domain with weakly labeled target domain, our PRE method bridges the sparse labels and global feature distributions in the target domain. Through dynamic expansion and rectification of pseudo-labels guided by the prototypes, PRE achieves promising land cover mapping results for 10 cities in different regions around the world, surpassing an overall accuracy of $80\%$. These experimental results demonstrate the ability of our approach to achieve cross-sensor, cross-category, and cross-continent domain adaptation for complicated categories using solely low-cost annotations. We believe that our work will stimulate further research in global high categorical resolution land cover mapping, ultimately enhancing Earth observation capabilities for environmental monitoring and sustainable development initiatives.

\bibliographystyle{IEEEtran}
\bibliography{reference}
\end{document}